\documentclass[10pt,journal,compsoc]{IEEEtran}
\usepackage{tabularx}
\usepackage{multirow} 
\usepackage{threeparttable}
\usepackage{float}
\usepackage{graphicx}
\usepackage{subfigure}
\usepackage{amssymb}
\usepackage{booktabs} 
\usepackage{siunitx} 
\usepackage{graphicx} 
\usepackage{times}
\usepackage{soul}
\usepackage{url}
\usepackage[hidelinks]{hyperref}
\usepackage[utf8]{inputenc}
\usepackage[small]{caption}
\usepackage{enumitem}
\usepackage{graphicx}
\usepackage{amsmath}
\usepackage{amsthm}
\usepackage{booktabs}
\usepackage{algorithm}
\usepackage{algorithmic}
\ifCLASSOPTIONcompsoc
  \usepackage[nocompress]{cite}
\else
  \usepackage{cite}
\fi
\ifCLASSINFOpdf
\else
\fi
\begin{document}
%
\title{Dual Latent State Learning: Exploiting Regional Network Similarities for QoS Prediction}
%
%
%
%

\author{Ziliang~Wang, Xiaohong~Zhang, Kechi Zhang, Ze Shi Li and Meng~Yan, Member, IEEE
\IEEEcompsocitemizethanks{


\IEEEcompsocthanksitem  

Ziliang Wang, Kechi Zhang are with Key Laboratory of High Confidence Software Technologies (Peking University), Ministry of Education; School of Computer Science, Peking University, Beijing, China\\
Xiaohong~Zhang, Meng~Yan are with Key Laboratory of Dependable Service Computing in Cyber Physical Society (Chongqing University),  Ministry of Education, China and School of Big Data and Software Engineering, Chongqing University, Chongqing 401331, China. \protect
\\
Ze Shi Li is in the Department of Computer Science at the University of Victoria, Canada
\IEEEcompsocthanksitem 
Xiaohong Zhang is the corresponding authors.\\
E-mail: xhongz@cqu.edu.cn
}
}
\IEEEtitleabstractindextext{%
\begin{abstract}
Individual objects (users or services) within a region usually have similar network states to each other because they usually come from the same region. 
Despite the similarity within regional networks, many existing techniques overlook this potential, thereby limiting the accuracy of Quality of Service (QoS) predictions.
In this paper, we introduce the regional-based dual latent state learning network(R2SL), a novel deep learning framework designed to overcome the pitfalls of traditional individual object-based prediction techniques for QoS prediction. 
R2SL first captures the regional network states by deriving two distinct regional network latent states: the physical-regional latent state and the virtual-regional latent state. 
These states are built using aggregated data from a physical area (such as the same city) or a virtual network area (such as the same service provider) rather than individual object data.
Then R2SL provides a specific Mixture of Experts(MOE) network based on sparse activation to realize the classification learning task of latent features of different regions.
Thirdly, R2SL adopts an enhanced smooth Huber loss function that adjusts Huber loss’s linear loss component to alleviate the impact caused by label imbalance. 
Finally, the perceptual network was used to interpret the integrated features, thereby realizing the QoS prediction.
Experimental results show that compared with the typical methods and the latest methods, the proposed method can obtain higher prediction accuracy, thus improving the research level of QoS prediction.
\end{abstract}

}

\maketitle

\IEEEdisplaynontitleabstractindextext

%
\IEEEpeerreviewmaketitle

\IEEEraisesectionheading{\section{Introduction}\label{sec:introduction}}
\IEEEPARstart{I}{n} recent times, various domains and applications, such as cloud services, online streaming, and e-commerce, are increasingly being offered as a service, resulting in a greater emphasis on ensuring optimal Quality of Service (QoS)~\cite{muslim2022s,zheng2020web}.
QoS values are frequently used as crucial inputs for various downstream service computing tasks, such as service recommendation~\cite{liu2019personalized,yao2014unified} and service composition ~\cite{gavvala2019qos,sefati2021qos}.
At the same time, some microservice optimization techniques require accurate predictive Quality of Service (QoS) models, such as autoscaling based on QoS~\cite{47,hussain2022new}.
Consequently, accurately predicting QoS has become a critical topic in service computing ~\cite{shao2007personalized,liu2019context,ghafouri2020survey}.
%
%
%
%
%

However accurately predicting QoS is significantly hindered by major challenges such as \textbf{data sparsity} and \textbf{label imbalance}, which affect the reliability and performance of service recommendations~\cite{zhang2018deep,14}.
Data sparsity includes both 1) the sparsity of QoS records for users and services, and 2) the sparsity of user and service features.
The sparsity in QoS records arises because, despite the vast number of web services in a recommendation system, users typically access only a small subset~\cite{zhang2019location}.
Feature sparsity results from the challenge of obtaining detailed network status information for users or service providers due to privacy restrictions and high data collection costs~\cite{86}.
Therefore, collaborative filtering (CF) has been widely adopted to mitigate the sparsity of QoS records by leveraging similarities among users or services~\cite{carlkadie1998empirical}.
While CF-based approaches attempt to alleviate data sparsity effects by identifying similar objects using available data, it often struggles to exploit contextual features.
To alleviate the effect of feature sparsity, several latent factors (LF) based QoS prediction approaches have been proposed~\cite{27,wu2022double,xu2021nfmf}.
For instance, Wang et al. used a latent state learning model to identify the network latent states of individual users and services~\cite{86,lu2023feature}.
Although these LF-based methods  alleviate the sparsity problem of data features by extracting latent features, the lack of QoS records remains a bottleneck that hinders the optimal performance of object-based latent state learning algorithms.
Particularly for individual users, who typically have only a few dozen QoS records, the sparse records make it challenging to learn their latent states.

The label imbalance problem arises from real-world service access scenarios: most records come from standard access procedures, while only a small portion results from abnormal accesses.
For example, although the service response time (RT) collected by Zhang et al. ranges from 0 to 20 seconds, more than 98\% of the data is concentrated in the range 0 to 5 seconds, leaving only of the data 2\% longer than 5 seconds.~\cite{3,zhang2021probability}.
Various remedies such as data augmentation, re-sampling, and enhanced robustness loss functions have been proposed to counter this issue\cite{chattopadhyay2022offdq,19}.
For instance, Zhang et al. investigated the appropriate coefficient for Huber loss to enhance model robustness~\cite{zhang2019location}.
%
%
However, these methods often overlook the distribution characteristics of QoS data, leading to impaired prediction accuracy due to the lack of targeted design.


\begin{figure}
\centerline{\includegraphics[scale=0.4]{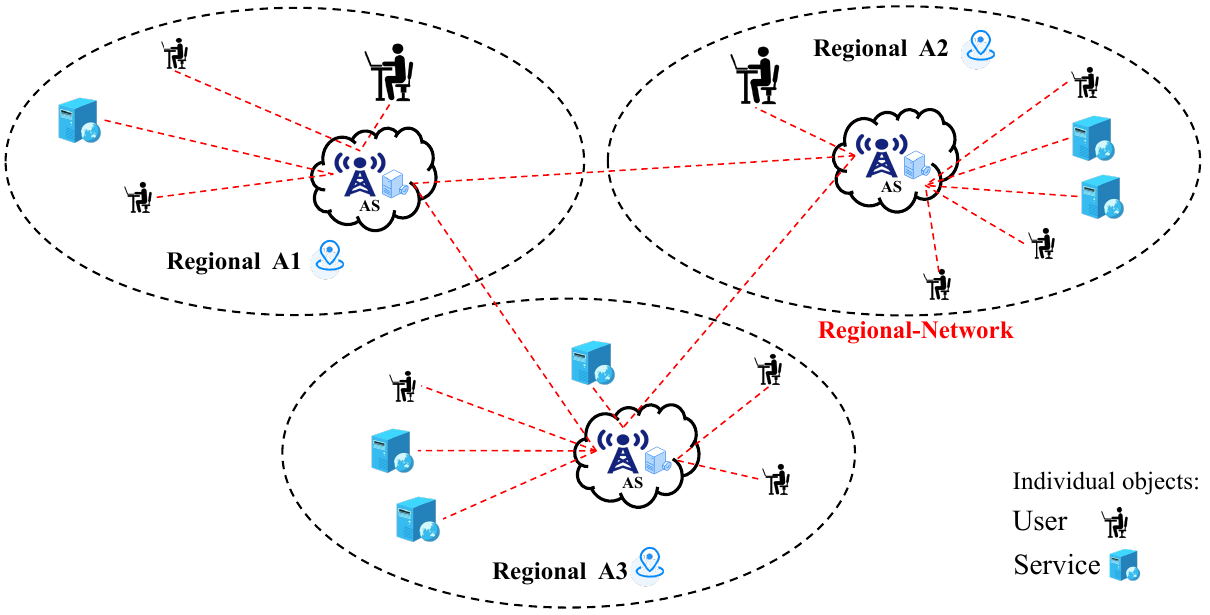}}
\caption{Objects typically exhibit network-state similarity within a region, especially if the regional network operates under the same autonomous system (AS).}
 \vspace{-1em}
\label{fig}
\end{figure}

To address these challenges, we propose the Regional Network Latent State Learning (R2SL) approach, which leverages regional network latent states to enhance the accuracy of QoS predictions.
R2SL introduces three innovative components for high-precision QoS prediction: a region-based latent state learning process, sparsely activated MOE for latent state awareness, and an enhanced smoothing Huber loss.
The first component is the region-based latent state learning process. Unlike traditional latent state algorithms that model user objects, R2SL models regions, with two default regional latent states: physical regional latent state and virtual regional latent state.
As shown in Figure 1, there are a large number of individual object data in the virtual network area and the physical area, which can effectively alleviate the problem of insufficient data volume.
The physical-network latent state derives latent states from QoS data of all objects within a specific city, while the virtual-network latent state learns latent states from QoS records within the same autonomous system (AS Code).
The second component is the sparsely activated MOE for latent state awareness. Based on a configurable number of regional latent states, we establish expert networks for each regional latent state feature space. 
To further mitigate the limitations of sparse data on model fitting, we implement a sparsely activated mechanism for feature screening tailored to each specific access task.
Finally, we deploy an enhanced smoothing Huber loss function, adjusting the linear loss component to mitigate label imbalances.

Our R2SL approach provides three groundbreaking advancements in QoS prediction.
1) Traditional latent state learning methods usually extract hidden features of a single object via the user-service interaction matrix.
These methods are often impacted by the lack of data for individual objects, such as users having only a dozen data points. R2SL addresses this by introducing regional latent states (physical and virtual) to replace object-specific data. 
2) R2SL introduces a sparsely activated MOE network for regional latent feature perception. 
Traditional latent state-aware networks usually utilize all the latent features to participate in the final QoS prediction task~\cite{86}.
In practical applications, there may be significant differences in the QoS features of different users accessing different services.
The sparse activation mechanism can effectively capture these differences, enhancing prediction accuracy.
For instance, in our model, when user 0 accesses service 47, the virtual latent features are universally activated compared to the physical latent features.
3) Recent approaches propose to use huber loss to achieve more robust QoS prediction models~\cite{zhang2019location}..
In QoS prediction tasks, the access time usually presents a long-tail distribution, and the traditional Huber loss is greatly affected by outliers.
 For instance, the predicted loss for normal access is 0.2, represented as 0.04 in Huber loss (squared loss), whereas the loss for an abnormal access could be 18 (linear part). This substantial loss can greatly impact the model.
By adjusting the coefficients of the linear part, the enhanced Huber loss provides a more balanced loss across different prediction errors, reducing the excessive weight of long-tail labels and improving overall prediction performance.

In summary, this paper's primary contributions are:
\begin{itemize}
\item [a)] We propose a latent discrete distribution-based algorithm to model regional latent states, alleviating the impact of data sparsity on the efficiency of traditional object-based latent state learning. 

\item [b)] We propose a fine-grained sparse-activated MOE network that constructs expert models of latent features in various regions. By dynamically adjusting the weights of the expert network output, the model can perform dynamic latent feature perception for different users and service access tasks. 

\item [c)] We analyze the distribution characteristics of QoS data and propose an enhanced Huber loss function tailored to its highly imbalanced nature. By refining the linear part of the Huber loss function, our method can better manage large errors and reduce the impact of outliers, thereby improving the overall accuracy and robustness of QoS prediction.

\item [d)] We conducted comprehensive experiments on QoS datasets collected from the real world. Experimental outcomes highlight R2SL's superior performance over existing state-of-the-art QoS prediction approaches.\footnote{Our replication package: The code base address will be updated after the paper is publicly available}
\end{itemize}

\section{Related work}\label{sec:Related work}
Current strategies for QoS prediction can be broadly divided into two categories: collaborative filtering-based methods~\cite{hussain2022new,muslim2022s,wu2020data,chowdhury2020cahphf,liu2019context,zheng2020web} and deep learning-based methods~\cite{liang2021recurrent,li2021topology,xia2021joint,ghafouri2020survey}.

\textbf{Collaborative Filtering-Based QoS prediction Approaches.} 
Collaborative Filtering (CF) is a widely used technique for QoS prediction, leveraging the similarity between objects to predict the QoS for a target request~\cite{1,2,3,4}.
 CF-based QoS prediction methods are generally classified into memory-based and model-based approaches. 
The memory-based methods use similarities among users or services to estimate QoS, whereas the model-based methods employ machine learning models to uncover the relationships between users, services, and QoS~\cite{13}.

Memory-based CF methods often use QoS metrics such as response time and network traffic, along with user and service attributes like location information, to determine similarities.
The core principle involves computing similarity scores relative to the target entity. These similarities can be calculated using user-based techniques (e.g., UPCC~\cite{12}), service-based methods (e.g., IPCC~\cite{13}), or combined user and service attributes (e.g., UIPCC~\cite{14}). While these strategies are simple, they are efficient since they utilize a single attribute to gauge similarity.
To improve prediction accuracy, Bellcore and colleagues incorporated contextual information, using evaluations from similar users to predict the target user's service assessment~\cite{79}.
In addition to memory-based methods, model-based techniques like matrix factorization (MF) are extensively used in QoS prediction~\cite{8,9}. 
Modern model-based methods enhance CF precision by incorporating contextual factors such as time and location. 
For instance, Zhang et al. proposed a time-aware framework that tailors QoS predictions to the specific timeline of the service user~\cite{35}.
Similarly, Wang and colleagues suggested a distance-based selection strategy that utilizes user coordinates~\cite{81}, while Chen et al. developed RegionKNN models that cluster users based on IP addresses and location similarity~\cite{2}. 
However, the high cost of data collection and concerns about user privacy limit the availability of such contextual information.  
Therefore, latent factor (LF)-based QoS predictors, such as Luo et al.'s nonnegativity constraint-based latent factor (NLF) model, are gaining popularity for their scalability and accuracy~\cite{29}. 

\textbf{Deep Learning-Based QoS prediction  Approaches.}
To enhance the nonlinear learning capability of collaborative filtering, He et al. introduced Neural network-based Collaborative Filtering (NCF), which replaces the traditional inner product with a neural architecture capable of learning arbitrary functions from data~\cite{20}.
Incorporating spatio-temporal data has also been a focus in QoS prediction. For example, Zhou et al. proposed a model where each time slice is equipped with a latent attribute to represent its state~\cite{83}. 
Similarly,  Wang et al. developed a motif-based dynamic Bayesian network that captures conditional dependencies over time for QoS prediction~\cite{wang2016online}. 
Xiong et al. proposed the Deep Hybrid Service Recommendation (DHSR) approach, which combines text similarity and an MLP network to uncover nonlinear relationships between services and mashups~\cite{xiong2018deep}. 
Additionally, to reveal hidden data, Wang et al. introduced a latent state model that uses latent factor algorithms to identify various latent attributes of users and services, thereby improving the prediction accuracy of deep neural network models~\cite{86}.
In the latest research, Resnet, graph convolutional network and other technologies are also widely used in QoS prediction~\cite{liu2023qosgnn}.


R2SL gleans insights from other realms, including models predicated on the object-based Latent state learning approaches~\cite{29} and the DL-based QoS prediction approach~\cite{86} techniques. 
%
A significant difference between R2SL and these models is that R2SL further improves the object-based latent state learning algorithm to apply on region objects and propose targeted perception networks.

\section{Approach}
The architecture of R2SL, depicted in Fig. 2, comprises four primary modules: a latent state learning module based on region objects, a gating network designed for specific access tasks, a sparse activation MOE network utilizing region state learning, and a QoS prediction network employing a multi-layer perception network.
\begin{figure*}
\centering
\includegraphics[width=0.88\textwidth]{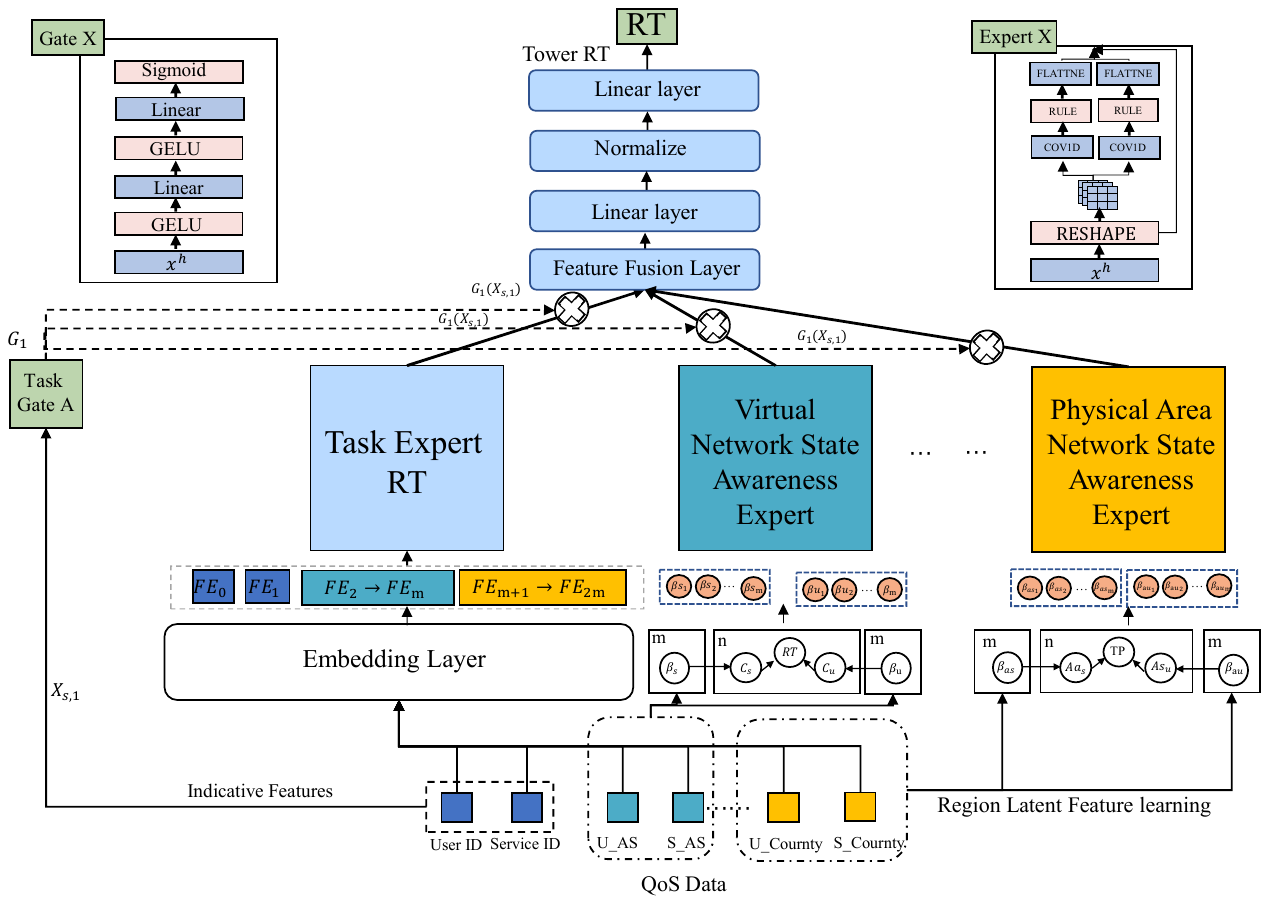}
\captionsetup{justification=centering}
 \vspace{-1em}
\caption{The R2SL method consists of four processes: a latent state learning module based on region objects, a gating network based on specific access tasks, a sparse activation MOE network based on region state learning, and a QoS prediction network based on multi-layer perceptron.}
 \vspace{-1em}
\end{figure*}

\subsection{Regional Network Latent State Learning}
\subsubsection{Latent Regional State Variable}

\begin{figure}[H]
\centerline{\includegraphics[scale=0.5]{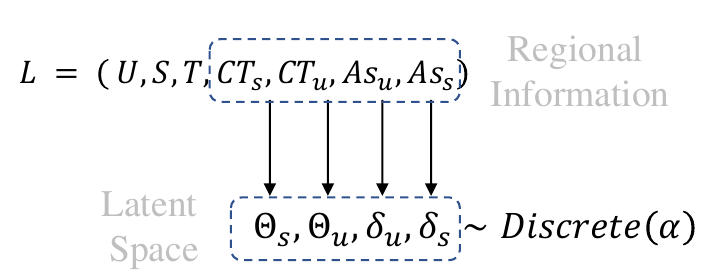}}
\caption{Different from traditional latent learning methods that model U and S, R2SL models the latent space of region objects.}
\label{fig}
\end{figure}
\indent Define QoS records  \( L = ( U, S, T, CT_s, As_s, CT_u, As_u) \), where \( U \) and \( S \) represent the user and service IDs, respectively. \( CT \) and \( As \) denote city code and AS code, respectively.
\( T \) signifies the QoS(RT or TP), such as response time for user \( U_i \) requesting \( S_i \), constrained to \( 0s < T < 20s \). 
City id and AS id for services are represented by \(  CT_s, As_s \), and by \(CT_u, As_u \) for users. The latent states set \( B = \{ b_k \} \) ( \( 1 < k < m \)) encompasses \( m \) latent state variables needed by the services' regional network (physical/virtual regional) and provided by the users' regional network (physical/virtual regional) to conduct activities.

To determine the latent distribution of regional states, we employ a latent state learning algorithm.
This approach aids in discerning the latent network states across diverse cities and AS. 
We initiate by defining four distributions, \( \Theta_{s} \) and \( \Theta_{u} \), symbolizing the virtual-network state frequencies for users and services.
Analogously, for the physical-network latent distribution, we have \( \delta_{s} \) and \( \delta_{u} \). Their mathematical representation is:
\begin{equation}
\centering
\Theta_{s},\Theta_{u},\delta_{u},\delta_{s}\sim Discrete(\alpha)
\end{equation}
In general, $\Theta_{s}$ and $\Theta_{u}$ represent the latent states of the physical area where users and services are located based on the city information, and $\delta_{u}$ and $\delta_{s}$ express the latent states of the virtual network area where users and services AS are located. 
Network data are typically limited and discrete. Service requests and response times, bandwidth, and other QoS metrics are often discrete values rather than continuous.
This makes using discrete distributions to model latent states more consistent with the characteristics of actual data. $\alpha$ is the random seed used to initialize the discrete distribution and defaults to $M$.

\subsubsection{Parameterized for QoS}
\indent To model the nuanced influence of various regional state frequencies and assignments on QoS, we identify primary contributing factors that are the statistical properties of QoS.
In our discussion, \( T_{i} \) represents the response time, although other metrics can be substituted. 
The observation value \( T \) is susceptible to changes in the network states. We assume that QoS is sampled from an exponential distribution. 
Based on the property of exponential distribution, it means that the smaller the value of quality of service, the value of the latent feature will be larger.
This assumption is derived from real-world experience, such as the response time is smaller when the network speed is faster.
Let \(\Phi\) represent the probability density function based on the exponential distribution:
\begin{equation}
\centering
\Phi (T_{i},\lambda _{i,j,k})=\lambda _{i,j,k}e^{(-T_{i}*\lambda _{i,j,k})}
\end{equation}
where
\begin{equation}
\centering
\lambda _{i,j,k}^{-1}=\begin{cases}
Cu_{j}Cs_{k}  & \text{ if }  T_{i}<\eta \\
Cu_{j}Cs_{k}  W & \text{ else }
\end{cases}
\end{equation}
Where \(0 < J, K < M\), the \(i\)-th value corresponds to the \(i\)-th QoS record.
\(M\) is the number of latent factors. 
Consider a scenario where each physical-virtual regional pair has been correctly assigned to a latent state. 
In this case, all service time values in the QoS log should precisely follow an exponential distribution characterized by the parameters \(C_u\), \(C_s\), and \(W\).
Intuitively, these factors influence the relationship between QoS and the assignments of physical/virtual latent states, thereby affecting the distribution of QoS values. 
When the QoS meets the desired threshold, such as a response time within 2.5 seconds, we model the QoS distribution using an exponential distribution with an expectation of \(C_u \cdot C_s\). 
Otherwise, we use an exponential distribution whose expectation is the product of all three parameters: \(C_u \cdot C_s \cdot W\). 
As shown in Figure 4, \(C_u\) and \(C_s\) are the complexity factors of the physical/virtual latent states, respectively, and \(W\) is the penalty factor, which defaults to 50.

\subsubsection{Physical/Virtual Latent State Sampling}
\indent \textbf{Physical-network state sampling:} This step considers the probability of assigning latent state $Z_u/T_u$ to user $U_i$ and assigning latent state $Z_s/T_s$ to service $S_i$ for each log record $x_i = (U_i, S_i, T_i)$. To establish the relationship between the latent state $Z_u$ and $U_i$, the user $u_i$ with physical area latent state $Z_u$ is sampled from the following distribution:
\begin{equation}
\centering
Z_{u}|\Theta _{u}\sim Discrete(\Theta _{u})
\end{equation}
In a similar vein, the probability of the latent state associated with the city \( CTs_{i} \)  in which the service location is expressed as \( Z_{s} \):
\begin{equation}
\centering
Z_{s}|\Theta _{s}\sim Discrete(\Theta _{s})
\end{equation}
\textbf{Virtual-network state sampling:} in the same way, the latent state $T_{s}$ of the AS gateway where the service is located is sampled:
\begin{equation}
\centering
T_{s} | \delta_{s} \sim Discrete(\delta_{s})
\end{equation}
For the user's virtual-network, we have:
\begin{equation}
\centering
T_{u} | \delta_{u} \sim Discrete(\delta_{u})
\end{equation}



\subsubsection{Quality of Service Sampling}
\indent The crucial step is sampling the response time $T_{i}$ given the i-th QoS record $L$. $T_{i}$ as the object of observation is affected by different factors of users and services. For each record, the conditional probability is used to represent its distribution:
\begin{equation}
\centering
T_{i}|(Z_{u},Z_{s})(T_{u},T_{s})(C_{u},C_{s}),W\sim \Phi (T_{i};\lambda _{i,j,k})
\end{equation}

where $\Phi$ is an exponential distribution whose expectation is defined in Section 3.1.2. In this formalization, a specific service is completed based on the complex factors $C_u$ and $C_s$, and the QoS metric $T_i$ is observed under the latent state conditions of the physical/virtual regions.

\begin{algorithm}
\caption{Network Latent State Learning Algorithm} 
\label{alg1} 

\begin{algorithmic}
\REQUIRE $L$,$n_*$ mean the number of *.\\
a : the prior parameter.\\
$\varrho$ : learning rate in GD.\\
m : the number of latent ability mixture components.
\ENSURE ($\Theta_u,\Theta_s,\beta_{u},\beta_{s}$)
\WHILE{$\tau^{t+1}/\tau ^{t}> \gamma $}
\STATE //E-step
\FOR{$i=1$ to $n$}
\FOR{$j=1$ to $m$}
\FOR{$k=1$ to $m$}
\STATE $G_{i,j,k}^{t}=\tau_{i,j,k}\Phi (T_i;\lambda _{i,j,k})$
\ENDFOR
\ENDFOR
\ENDFOR
\STATE //M-step
\\
\FOR{$i=1$ to $m$}

\FOR{$q=1$ to $n_{CTu}$}
\STATE $\Theta_{u\left \{ i,q \right \}}^{t+1}=\sum _{i^{'}=1}^{n}\sum _{k=1}^{m}G_{i^{'},i,k}^{t}I(CT_{u_{i}}=q)$
\ENDFOR
\FOR{$p=1$ to $n_{CTs}$}
\STATE $\Theta_{s\left \{ i,q \right \}}^{t+1}=\sum _{i^{'}=1}^{n}\sum _{j=1}^{m}G_{i^{'},j,i}^{t}I(CT_{s_{i}}=p)$
\ENDFOR

\FOR{$q=1$ to $n_{ASu}$}
\STATE $\beta_{u\left \{ i,q \right \}}^{t+1}=\sum _{i^{'}=1}^{n}\sum _{k=1}^{m}G_{i^{'},i,k}^{t}I(AS_{u_{i}}=q)$
\ENDFOR
\FOR{$p=1$ to $n_{ASs}$}
\STATE $\beta_{s\left \{ i,q \right \}}^{t+1}=\sum _{i^{'}=1}^{n}\sum _{j=1}^{m}G_{i^{'},j,i}^{t}I(AS_{s_{i}}=p)$
\ENDFOR
\\

\ENDFOR
\\
\STATE  //GD-step
\FOR{$q=1$ to $n$}
\STATE $Cu_{i}^{t+1}=Cu_{i}^{t}+\varrho *\frac{\partial P(\Gamma |L)^{t}}{\partial Cu_{i}^{t}}$
\ENDFOR
\FOR{$q=1$ to $n$}
\STATE $Cs_{i}^{t+1}=Cs_{i}^{t}+\varrho *\frac{\partial P(\Gamma |L)^{t}}{\partial Cs_{i}^{t}}$
\ENDFOR \\
\STATE$t=t+1$
\STATE$W^{t+1}=W^{t}+\varrho*\frac{\partial P(\Gamma |L)^{t}}{\partial W^{t}}$\\
\STATE$\tau^{t+1}= P(\Gamma |L)$

\ENDWHILE
\\
return ($\Theta_u,\Theta_s,\beta_{u},\beta_{s}$)
\end{algorithmic}
\end{algorithm}

\subsubsection{Parameter estimation}
\begin{figure*}
\centerline{\includegraphics[scale=0.65]{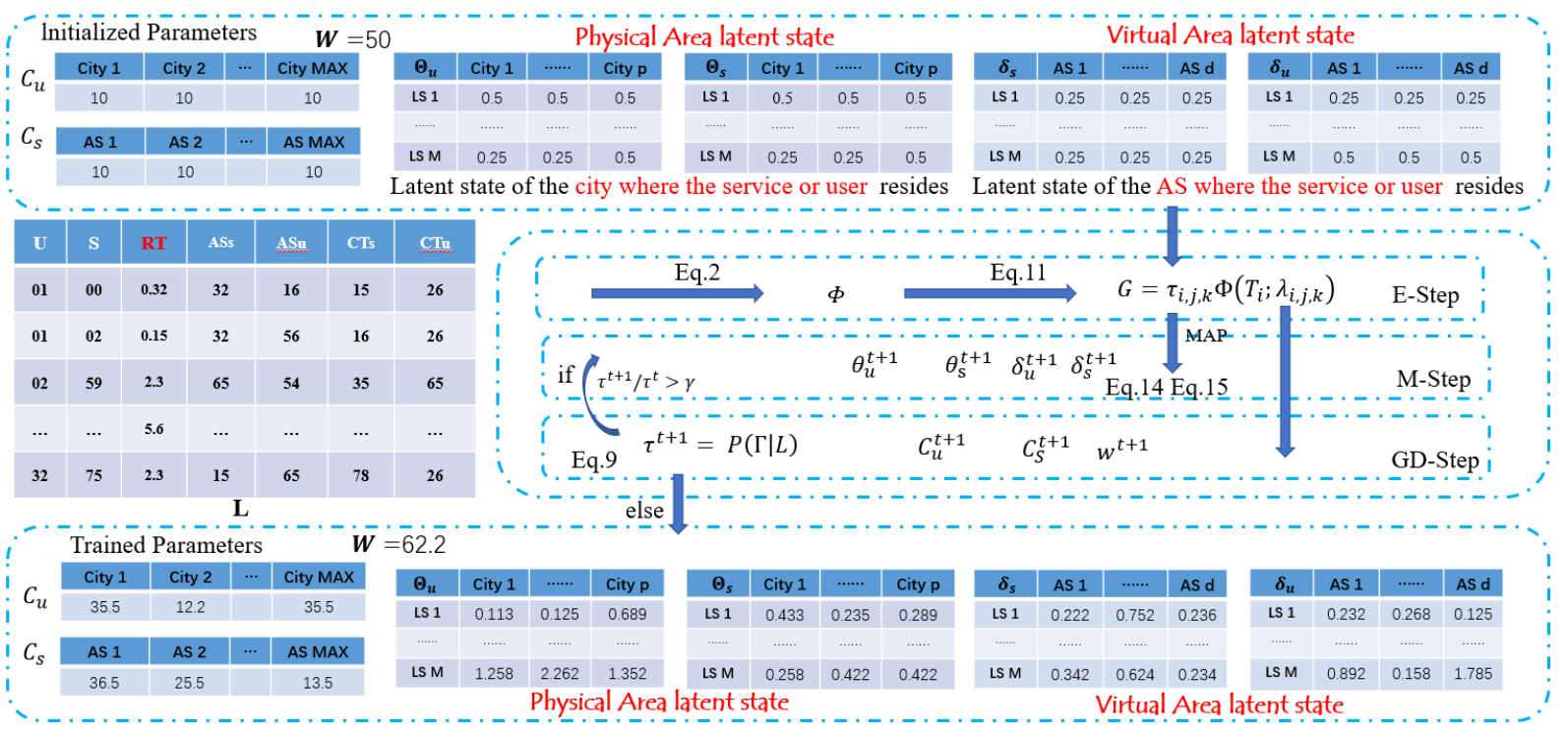}}
\caption{The graphical model of regional latent state learning process}
 \vspace{-1em}
\label{fig}
\end{figure*}
The $\Lambda  (\Theta _{u},\Theta _{s},\delta_{u},\delta_{s},W,C_{u},C_{s})$ is the set of parameters for regional network latent state. 
To complete the parameter estimation process, R2SL uses the maximum a posteriori (MAP) algorithm.
The distribution $P(\Lambda  |L)$ is learned by the true QoS records $L$.
The calculation process is as follows:
\begin{equation}
\centering
P(\Lambda |L)=\prod_{i=1}^{n}\sum_{j=1}^{m}\sum_{k=1}^{m}\Phi (T_{i};\lambda _{i,j,k})\tau _{i,j,k}
\end{equation}
where
\begin{small}
\begin{equation}
\centering
\tau _{i,j,k}=\delta_{u\left \{ j,As_{u_i} \right \}}\delta_{s,\left \{ k,As_{s_i} \right \}}\Theta _{u\left \{ j,CT_{u_i}\right \}}\Theta _{s\left \{ k ,CT_{s_i}\right \}}
\end{equation}
\end{small}
Eq. (9) is the product of $n$ QoS records in L, $m$ states-to-network latent state assignment probabilities for each of the users physical/virtual latent state and $m$ states-to-network assignment probabilities for each of the services physical/virtual latent state.

To solve this MAP (Maximum A Posteriori) problem, we employ the expectation-maximization (EM) algorithm for the subset of parameters $(\delta_u, \delta_s, \Theta_u, \Theta_s)$, and gradient descent (GD) for the parameters $(C_u, C_s, W)$ simultaneously.


The E-step refers to the expectation step, in which we calculate the conditional distribution of the physical/virtual latent state to user assignment probability $u_i$ and the physical/virtual latent state to service assignment probability $s_i$ using Bayes' theorem, given the current estimation of parameters $\Gamma$:
\begin{equation}
\begin{aligned}
G_{i,j,k}^{t}&=P(CT_{u_{j}},CT_{s_{k}},AS_{u_j},AS_{s_k}|u_{i},s_{i},t_{i},\Gamma )\\
&=\tau _{i,j,k}\Phi (T_{ij};\lambda _{i,j,k})
\end{aligned}
\end{equation}
We use maximizing the condition expectation to update parameters $\Gamma$.
\begin{small}
\begin{equation}
\begin{aligned}
T(\Gamma |\Gamma ^{t})&=\sum_{i=1}^{n}\sum_{j=1}^{m}\sum_{k=1}^{m}log(P(\Gamma |Q))G_{i,j,k}^{t}\\
&=\sum_{i=1}^{n}\sum_{j=1}^{m}\sum_{k=1}^{m}log(\tau _{i,j,k}\Phi (T_{ij};\lambda _{i,j,k}))G_{i,j,k}^{t}
\end{aligned}
\end{equation}
\end{small}
The M-step refers to the maximization step, in which we update parameters by maximizing the conditional expectation $ T(\Gamma \mid \Gamma^{t})$.

\begin{equation}
\begin{aligned}
\Theta_{s}^{t+1},\Theta_{u}^{t+1},\delta _{s}^{t+1},\delta _{u}^{t+1}=maxT(\Gamma |\Gamma ^{t})
\end{aligned}
\end{equation}


We revise $\Theta_{u}^{t+1}$, a matrix within the space $\mathbb{R}^{m \times n_{CT_u}}$, where $n_{CT_u}$ denotes the number of cities where the user is located. Here, $\Theta_{u\{k,p\}}$ signifies the probability that city $CT_{u_i}=p$ possesses latent state $k$.
\begin{equation}
\begin{aligned}
\Theta_u^{t+1} &= \arg \max_{\Theta_u} \mathbb{E}_{\Theta_u}[\log P(\Lambda | L, \Theta_u)]\\
&=\sum_{i=1}^{n}\sum _{k=1}^{m}G_{i,j,k}^{t}I(CT_{u_{i}}=q)
\end{aligned}
\end{equation}
Similarly, for $\Theta_{s}$:
\begin{equation}
\Theta_{s_{k,q}}^{t+1}=\sum _{i=1}^{n}\sum _{j=1}^{m}G_{i,j,k}^{t}I(CT_{s_{i}}=p)
\end{equation}

where $I(CT_{u_{i}}=q)$ means that the final value of the function is 1 when user's city code is equal to q and 0 otherwise; $I(CT_{s_{i}}=q)$ means that the value of the function is 1 when service's city code is equal to p and 0 otherwise.
Finally, we obtain the user virtual-network latent state distribution $\delta_u$ and the service virtual-network latent state distribution $\delta_s$  through Algorithm 1.

In the GD-step, we determine the parameters $c_s$, $c_u$, and $w$ using gradient descent (GD). Following the M-step, these parameters are adjusted by moving in the direction of their gradients with a learning rate $\varrho$. The detailed calculation procedure is presented in Algorithm 1.

\subsection{Sparsely Activated Latent State-aware Networks}

\textbf{Feature Embedding.} For deep neural networks to effectively learn prominent features, we input all features such as $L$ and regional latent fetaures into the embedding layer provided by TensorFlow. Conceptually, the embedding layer can be regarded as a linear layer where the bias is 0 ~\cite{1}. Using this approach, the distinguishable features (i.e., ID, city code, and AS code) are mapped into distinct vectors.
The dimensions of this feature are defined as:
\begin{equation}
\centering
x_{i}^{'}=f(W_b \cdot x_i+b_i).
\end{equation}
where $f$ are feature embedding layer.  $x_i$ includes all the known features in $L$.


\noindent\textbf{Regional Expert Network.} 
The R2SL network has two different types of expert networks: physical area network state awareness experts and virtual state awareness experts. The number of experts can be adjusted according to the number of features in the region.
The first are task-dependent task expert models that accept features from a specific task as input.
The second are domain experts who accept as input features from the same domain for different tasks.
Both types of experts share the same structural features. The specific process is as follows:

\begin{equation}\label{pri}
E(x_i)=R(f(x_{i}^{'},\Theta_u^{'},\Theta_s^{'},\beta_{u}^{'},\beta_{s}^{'})).
\end{equation}
where $f$ is the feature fusion process, where different expert networks use different feature vectors as input. As an example, the input of task expert RT is the known feature vector $F_{0-k}$, physical area latent feature vector $F_{k-m}$ and  virtual network latent feature $F_{m-2m}$ . The domain expert only accepts feature vectors in the data of different regions. $R$ add dimension to the fused features to fit the input of the convolution operation.
\begin{equation}\label{pri}
E_t=f_{cov}^{[k]}(E(x_i)).
\end{equation}
where $f_{cov}^{[k]}$ is a multi-scale convolutional perception network with convolution kernel $3 \times 1,5\times 1$ to obtain two output features. 

\begin{equation}\label{pri}
E_i=w_{i}^{out}\cdot GELU(E_t).
\end{equation}
The output of the multi-scale convolutional network is entered into the GELU (Gaussian error linear unit) activation function and fused to obtain the output $E_i$ of the expert network.

\noindent\textbf{ Latent features-aware Sparse Gate to Experts.}  The R2SL network proposes latent features-aware gating functions that enable it to identify different regional latent features and thus route the output of a specific expert to the final task Tower network.

For each particular access task, that is, a particular user accessing a particular service, there is its own trainable weight matrix $W_{g,t}$. The dimension is $R^{N\times H}$, Where $N$ is the number of experts and $H$ is the dimension of the hidden states.
\begin{equation}\label{pri}
g_t=Softmax(f^{line}(u_i,s_i)).
\end{equation}
where the inputs to the gating network are the user id vector $u_i$ and the service id vector$s_i$. The structure of the gated network is shown in Figure 2, activated using Sigmoid function after two linear layers.




\noindent\textbf{Prediction network.} 
The input of the final prediction network is the fusion of the outputs of task experts and domain experts. The fusion process is as follows:
\begin{equation}\label{pri}
h_x=C(g_t*E_1,...,g_t*E_i)
\end{equation}
where $c$ is the concatenate operation, $g_t$ is the gated network output, and $E_i$ is the expert network output.

The decoder composed of another nonlinear layer completes the Qos prediction. The specific process is as follows:
\begin{equation}
\hat{y}=f^{[2^{v},2^{v-1},2^{v-2},1]}(h_x).
\end{equation}

where $f^{[2^{v},2^{v-1},2^{v-2},1]}$ denotes the fully connected layer, $V$ is the count of neurons in this layer, and $\hat{y}$ the prediction value. The default V is set to 5 in this paper.

\subsection{S-Huber Loss Function}

For QoS prediction, the Mean Absolute Error (MAE) loss function is prevalently employed~\cite{86,zhang2018deep}. The MAE loss function directs the model to concentrate on normal values, while being minimally affected by outliers. Some research has adopted the Huber loss to enhance the model's attention to outliers~\cite{zhang2019location}. Broadly speaking, the Huber loss demonstrates higher sensitivity to outliers compared to the MAE loss and exhibits greater robustness than the Root Mean Square Error (RMSE) loss.

\begin{figure}
\centerline{\includegraphics[scale=0.3]{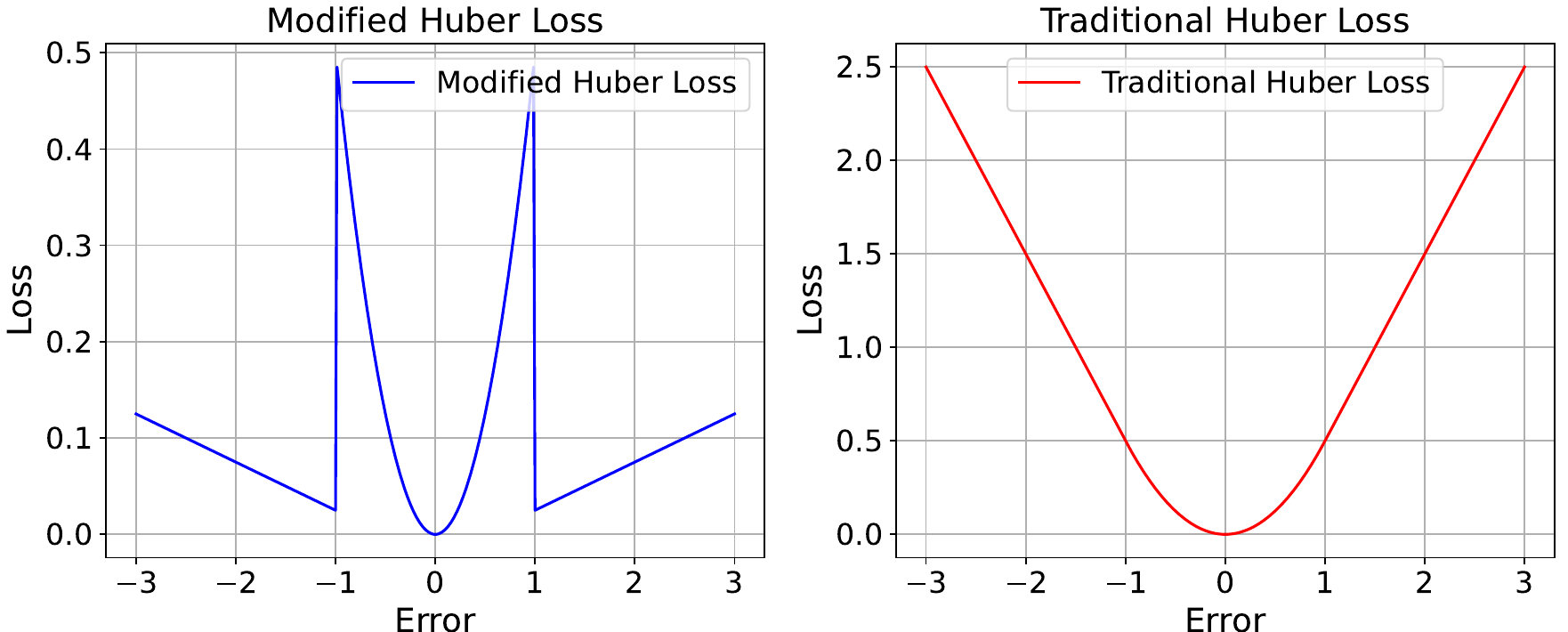}}
\caption{Errors with absolute errors greater than one are scaled down(fig.5 a). This is because the small error of QoS prediction is much smaller than the error caused by outliers after being squared, which makes the model training not smooth enough and difficult to converge. }
 \vspace{-1em}
\label{fig}
\end{figure}

In our exploration of QoS data distribution detailed in Sec. 4, we discerned that the linear component of the Huber loss remains significantly large, causing the model to still not adequately focus on outliers. R2SL addresses this by re-weighting the linear component of the Huber loss function in accordance with the intrinsic characteristics of the QoS data distribution. The new loss function, called S-Huber loss, alleviates the problem that the long-tail labels of QoS data have too much influence on model learning.

The S-Huber loss function between \( y \) and \( \hat{y} \) is given by:

\begin{equation}
S-H(y,\hat{y})=\left\{\begin{matrix}
\frac{1}{2}(y-\hat{y})^2 & if \ |y-\hat{y}|_{abs}<\varsigma,\\ 
\psi (\varsigma |y-\hat{y}|_{abs}-\frac{1}{2} \varsigma ^2)& otherwise.
\end{matrix}\right.
\end{equation}

Here, \( y \) represents the actual label, \( \hat{y} \) is the predicted value, and \( \varsigma \) is the Huber loss hyperparameter, defaulting to 0.5.

\noindent\textbf{Motivation and Improvement.}
The Huber loss function boasts enhanced resilience to outliers compared to both MAE and MSE loss functions. By definition, the Huber loss equates to RMSE when the error is smaller than \( \varsigma \).
Conversely, for errors surpassing \( \varsigma \), the loss corresponds to MAE. However, extreme values in QoS data render the Huber loss suboptimal for directing model training.
For instance, a long-tail label (e.g., 20s) might yield a linear loss exceeding 15, while a standard label with a smaller prediction error (e.g., 0.5s) might result in an MSE loss of only 0.25. 
This pronounced disparity causes the model to be disproportionately swayed by long-tail labels. To mitigate this, we introduced a weighting factor to the linear loss segment. 
As shown in Figure 5, adjusting this coefficient's value allows the model to better characterize the long-tail labels. Therefore, diverging from the traditional Huber loss function, \( \psi \) acts as the weight for the linear loss component and is set at 0.05 in this paper.

\section{STUDY SETUP}\label{sec:STUDY SETUP}

\begin{table}
\centering
\scriptsize
\caption{Division of all designed dataset cases.}
\begin{tabularx}{8cm}{lllXXX}
\hline

\hline
No.&Density&\scriptsize{Train:Test:Valid}&\scriptsize{Train}&Test&\scriptsize{Validation}\\
\hline
RT:D1.1&0.02 & 2\%:78\%:20\% & 37,375 &1,310,535 &369,638\\
RT:D1.2&0.04 & 4\%:76\%:20\% & 74,969&1,572,292&369,638\\
RT:D1.3&0.06& 6\%:74\% :20\%& 112,016  &1,206,517&369,638 \\
RT:D1.4&0.08 & 8\%:72\% :20\%& 150,071 &1,172,461  &369,638\\
RT:D1.5&0.10 & 10\%:70\% :20\%& 186,059  &1,140,269& 369,638\\
\hline
TP:D2.1&0.02 & 2\%:78\%:20\% & 33,034 &1,288,340 &330,343\\
TP:D2.2&0.04 & 4\%:76\%:20\% &66,068&1,255,305&330,343\\
TP:D2.3&0.06 & 6\%:74\% :20\%& 99,103  &1,222,271&330,343 \\
TP:D2.4&0.08 & 8\%:72\% :20\%& 132,137 &1,189,236 &330,343\\
TP:D2.5&0.10 & 10\%:70\% :20\%& 165,171  &1,156,202&330,343\\
\hline
\end{tabularx}
\end{table}

\begin{figure} 
\centering 

\subfigtopskip=2pt 
\subfigbottomskip=2pt 

\subfigure[QoS data distribution]{\includegraphics[width=0.49\linewidth]{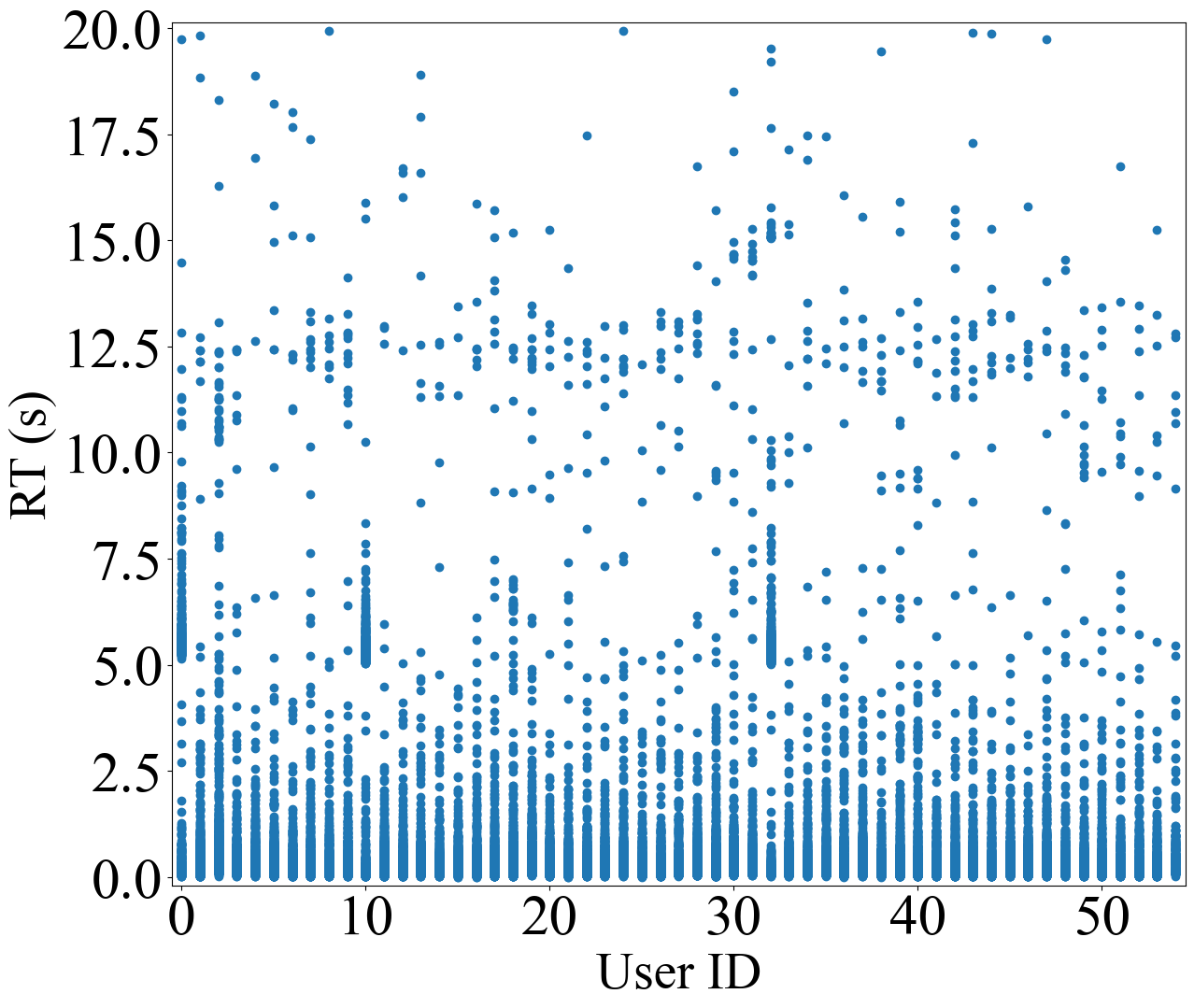}}
\subfigure[Data distribution of five user]{\includegraphics[width=0.49\linewidth]{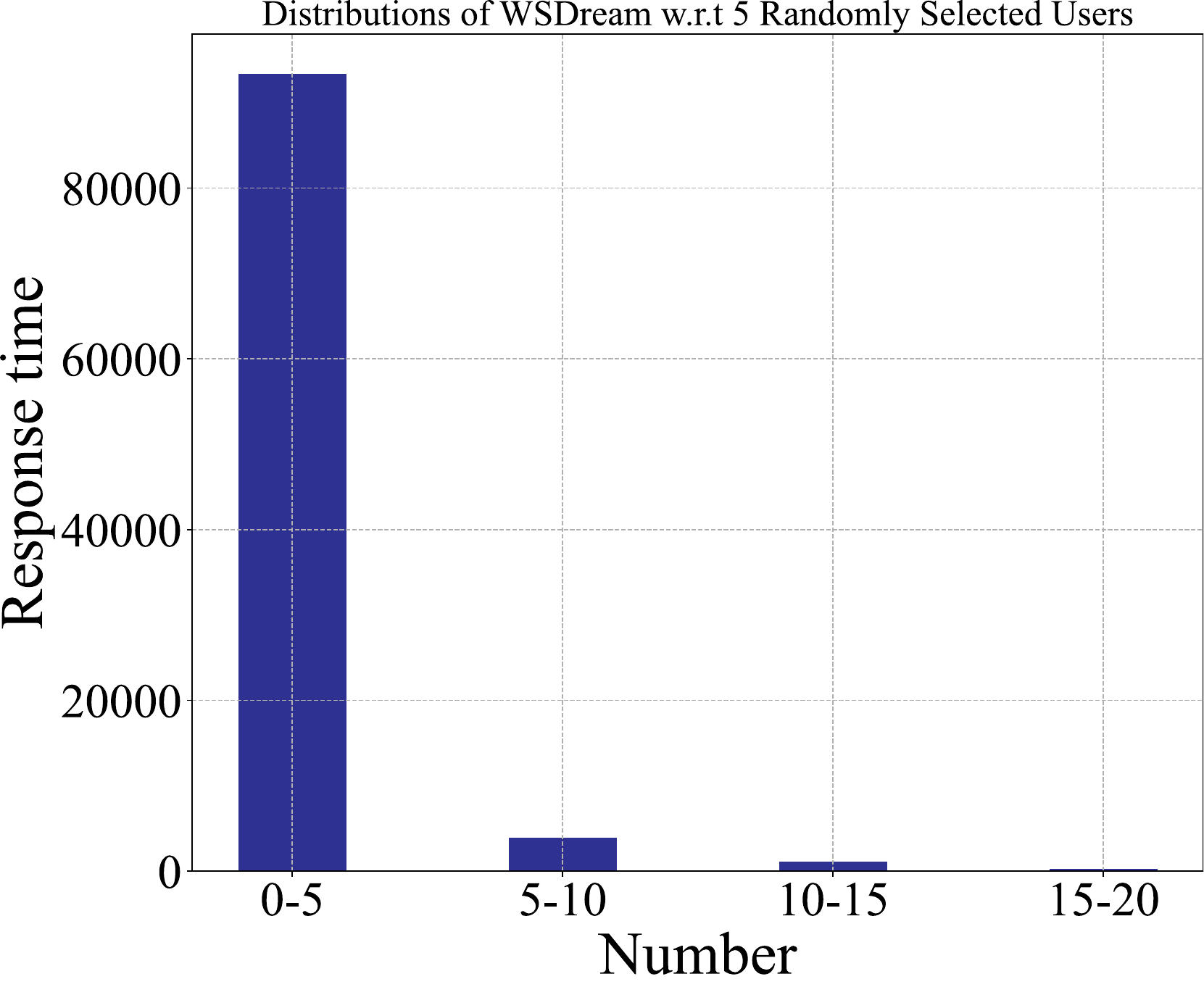}}
\subfigure[A user accesses service 0-1000]{\includegraphics[width=0.49\linewidth]{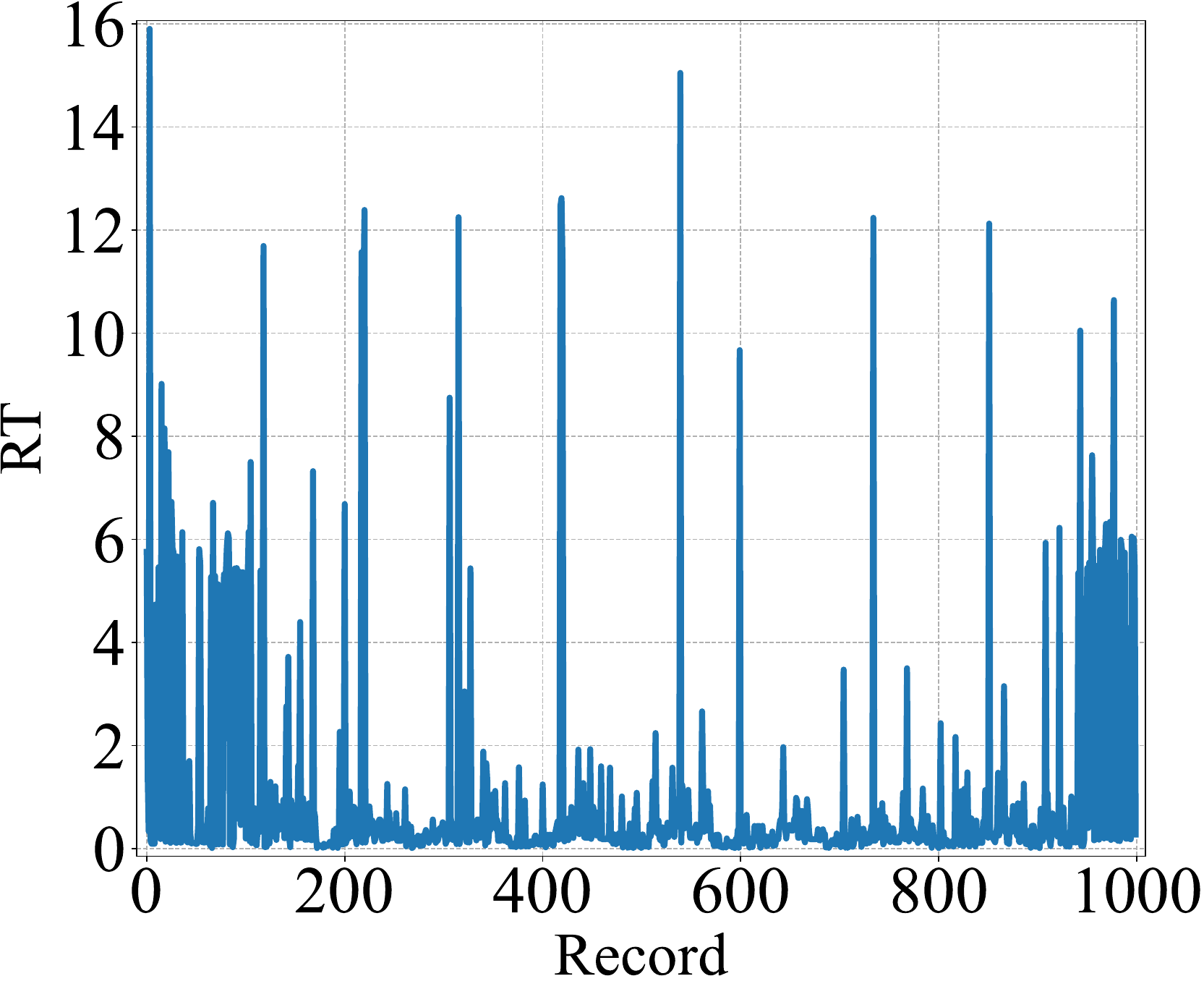}}
\subfigure[User 0-350 accesses the service A]{\includegraphics[width=0.49\linewidth]{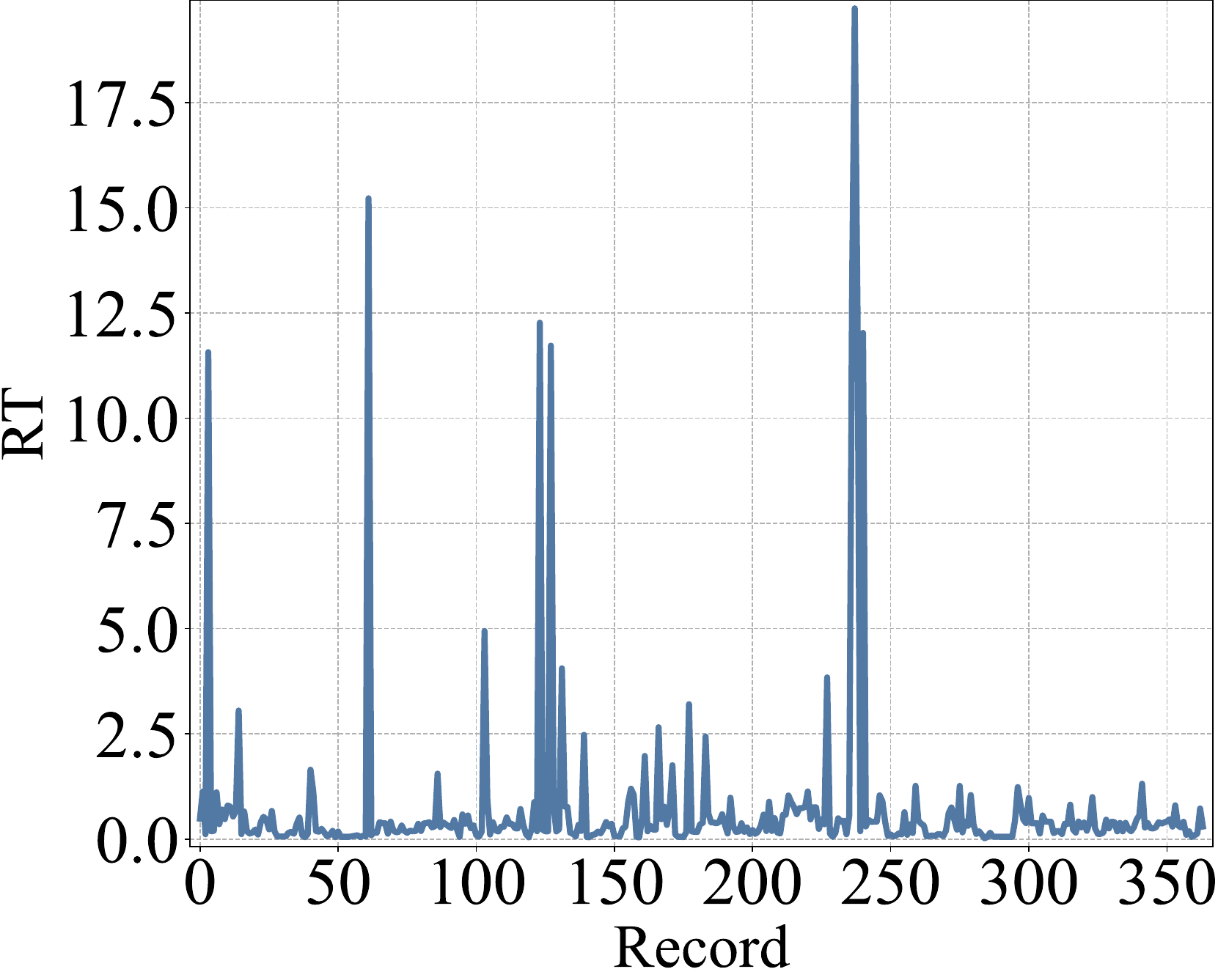}}

\vspace{-1em}
\caption{Data distribution of QoS datasets}
 \vspace{-1em}
\end{figure}


\subsection{Datasets}

We conducted validation experiments on publicly available benchmark datasets. The QoS dataset, termed WS-Dream, contains service data harvested from real-world web systems as detailed by ~\cite{3}. The WS-Dream dataset involves service records of 339 users accessing more than 5,000 services and contains response times for more than 1.9 million web service requests. WSDream contains two kinds of QoS attributes,i.e., response time (D1) and throughput (D2).

Of significance, the dataset utilized in this research is derived from the most recent investigation by ~\cite{77}, which refines the WS-Dream data by removing outliers. Adhering to their methodology, we employed the iForest (isolation forest) approach for outlier detection, maintaining the detection parameters consistent with ~\cite{77}. The threshold for outlier scoring is in alignment with ~\cite{77}, set at 0.1. The dataset is as follows:


For a comprehensive analysis, the dataset was segmented into five divisions, as presented in Tab. 1. These divisions were made to emulate data sparsity scenarios in real-world production environments and to ensure robust comparative experimentation. The data distribution for the WS-Dream dataset is depicted in Fig.3. A conspicuous observation from this figure is the pronounced label imbalance within the QoS dataset.

\begin{table*}
\begin{threeparttable}
  \centering

\selectfont
  \caption{PERFORMANCE COMPARISON OF QOS PREDICTION MODELS ON RESPONSE TIME}
  \label{tab:performance_comparison}
\setlength{\tabcolsep}{4.5mm}{
    \begin{tabular}{|c|c|c|c|c|c|c|c|c|c|c|}
    \hline

   \multirow{2}{*}{\textbf{Method}} &\multicolumn{2}{c|}{\textbf{D1.1}}&\multicolumn{2}{c|}{\textbf{D1.2}}&\multicolumn{2}{c|}{\textbf{D1.3}}&\multicolumn{2}{c|}{\textbf{D1.4}}&\multicolumn{2}{c|}{\textbf{D1.5}}\cr \cline{2-11}
&MAE   &RMSE   &MAE   &RMSE   &MAE   &RMSE   &MAE  &RMSE &MAE  &RMSE \cr
    \hline
 \hline
 UPCC&0.542 &1.022 &0.466 &0.820 &0.428 &0.787&0.389&0.754&0.555&1.317\cr
D2E-LF&0.653&1.638 &0.633&1.577 &0.607 &1.564&0.600&1.563&0.590&1.556\cr

NFMF&0.205 &0.543 &0.142 & 0.450&0.131 & 0.423&0.119 &0.406 &0.115& 0.409 \cr
 LDCF&0.349 &0.987 &0.279 &0.794 &0.247 &0.751 &0.240 &0.711 &0.213 &0.692 \cr
 CMF&0.294 &0.510&0.244&0.463 &0.207 &0.413 &0.186&0.390 &0.169 &0.367 \cr
   NCRL& 0.263&0.632 &0.252 & 0.772 &0.221 &0.722 & 0.201& 0.662 &0.182&0.660 \cr
QoSGNN&0.246 &0.467&0.240 &0.394 &0.191 &0.364&0.188&0.368&0.168 &0.355 \cr \hline
  \textbf{R2SL}&\textbf{0.151} &\textbf{0.440} &\textbf{0.124} &\textbf{0.393} &\textbf{0.114} &\textbf{0.359} &\textbf{0.109} &\textbf{0.342} &\textbf{0.104} &\textbf{0.335} \cr
    \hline

\end{tabular}  }
\end{threeparttable}
\end{table*}

\begin{table*}
\begin{threeparttable}
  \centering

\selectfont
  \caption{PERFORMANCE COMPARISON OF QOS PREDICTION MODELS ON THROUGHPUT}
  \label{tab:performance_comparison}
\setlength{\tabcolsep}{4.5mm}{
    \begin{tabular}{|c|c|c|c|c|c|c|c|c|c|c|}
    \hline

   \multirow{2}{*}{\textbf{Method}} &\multicolumn{2}{c|}{\textbf{D2.1}}&\multicolumn{2}{c|}{\textbf{D2.2}}&\multicolumn{2}{c|}{\textbf{D2.3}}&\multicolumn{2}{c|}{\textbf{D2.4}}&\multicolumn{2}{c|}{\textbf{D2.5}}\cr \cline{2-11}
&MAE   &RMSE   &MAE   &RMSE   &MAE   &RMSE   &MAE  &RMSE &MAE  &RMSE \cr
    \hline
 \hline
 UPCC&28.31 &60.12 &20.25 &52.58 &19.32 &48.45&18.76&45.54&15.17&42.69\cr
D2E-LF&18.30&27.06 &12.85&25.20 &11.82 &24.29&11.28&20.45&11.29&20.01\cr
 CMF&18.39 &41.98&15.71 &39.36 &14.32 &37.857 &13.32 &36.64 &12.73 &36.64 \cr
NFMF&16.23 &35.44 &11.25 & 29.50&8.78 & 26.54&7.68 &26.21&7.32& 25.65 \cr

 LDCF&14.56 &35.45 &9.78 &28.80&9.04 &27.57 &8.96&27.64 &8.66 &27.20 \cr
   NCRL& 12.10&31.94 &8.75 & 25.44&8.18 &24.44 & 7.17& 22.66 &7.47&23.11 \cr

QoSGNN&14.95 &35.52&10.98 &30.58 &9.56&28.10&9.00&27.00& 7.73 &25.31 \cr \hline

  \textbf{R2SL}&\textbf{9.28} &\textbf{25.91} &\textbf{7.86} &\textbf{22.84} &\textbf{7.31} &\textbf{21.83} &\textbf{6.85} &\textbf{20.91} &\textbf{6.65} &\textbf{20.27} \cr
    \hline

\end{tabular}  }
\end{threeparttable}
 \vspace{-1.5em}
\end{table*}
 



Fig.3(a) delineates the QoS records of 50 users, chosen randomly from the WS-Dream dataset, constituting 277,615 QoS record values. The data exhibits a mean of 0.770 and a variance of 3.454. Delving deeper, 95.10\% of the labels register values below 5, while the remaining 4.90\% exceed this value. Fig.3(b) catalogues the request records of 5825 services sourced from five users. It's evident that the response time for a preponderant number of requests is under \(5s\).

Fig.3(c) showcases the response time for user A when accessing services ranging from 0 to 1000. Predominantly, user A achieves access within \(4s\), yet there are certain services where the response time overshoots \(10s\). Conversely, Fig.3(d) demonstrates that for Service S, while the majority of user response times hover below \(2.5s\), some instances report times exceeding \(10s\). This underscores the influence of both user and service network states on the resultant quality of web requests.

\subsection{Evaluation Metrics}
The accuracy of the QoS prediction is an important criterion for evaluating the performance of the model.
Two metrics (i.e. MAE and RMSE) are commonly used to measure accuracy in most QoS prediction studies~\cite{77}.
The mean absolute error function (MAE) is defined as follows:
\begin{equation}
\centering
MAE=\frac{\left ( \sum_{i,j}^{} \left |r_{i,j}-\hat{r}_{i,j} \right |_{abs}\right )}{ N }
\end{equation}
where $r_{i,j}$ is the real QoS value (e.g., RT or TP) and $\hat{r}_{i,j}$ is the predicted value from predictive models.
The root mean squared error (RMSE) is defined as follows
\begin{equation}
\centering
RMSE=\sqrt{\tfrac{\left ( \sum_{i,j}^{} (r_{i,j}-\hat{r}_{i,j} )^{2}\right ) }{ N }}
\end{equation}
where N is the count of records.
For either the MAE indicator or the RMSE indicator, a lower value means higher predictive accuracy.

\subsection{Baseline Methods}

We compare our R2SL approach with the following these methods:


\par UPCC~\cite{13}: UPCC uses similar behavioural information from different users to achieve QoS prediction through collaborative filtering.

\par D2E-LF~\cite{wu2022double}: D2E-LF employs both inner product space and distance space to model LFA-based QoS predictors. It utilizes both L1 and L2 norm-oriented loss functions, and combines the predictions from these models using a weighting strategy to achieve higher accuracy in QoS prediction.

\par NFMF~\cite{xu2021nfmf}: NFMF, or Neural Fusion Matrix Factorization, combines neural networks and matrix factorization to perform non-linear collaborative filtering for latent feature vectors of users and services. It also considers context bias and employs multi-task learning to reduce prediction error and improve the predicted performance.

\par LDCF~\cite{21}: LDCF maps location features into high-dimensional dense embedding vectors and uses a multilayer perceptron (MLP) to capture high-dimensional and nonlinear characteristics. It employs a loss function designed to handle data sparsity and embeds a similarity adaptive corrector (AC) in the output layer to improve predictive quality.
\par CMF~\cite{77}: CMF utilizes Cauchy loss to measure the discrepancy between observed and predicted QoS values, making it resilient to outliers. It also incorporates temporal information to provide time-aware QoS predictions.
\par NCRL~\cite{zou2022ncrl}: NCRL employs a location-aware two-tower deep residual network to extract latent features of users and services for neural QoS prediction. It integrates these features to generate similar neighborhoods for collaborative prediction based on historical QoS data, enabling adaptive and accurate QoS prediction.

\par QoSGNN~\cite{liu2023qosgnn}: QoSGNN leverages Graph Neural Networks (GNNs) to jointly model the interactions between users and services and learn better feature embeddings. This framework systematically applies GNN principles to QoS prediction, offering improvements in prediction performance, cold start, scalability, robustness, and fairness.

In all experimental results, each method will be run five times and the results will be averaged for a fair comparison and other settings remain the same as CMF~\cite{77}.






\vspace{-0.5em}
\section{EXPERIMENTS}\label{sec:EXPERIMENTS}

\subsection{Prediction Performance Comparison}

In this section, we compare the prediction performance of different QoS prediction models on response time (RT) and throughput (TP). 
The experimental results are summarized in Tables 2 and 3. We report the mean and standard deviation of performance metrics (MAE and RMSE) from five independent experiments with different seeds. Based on these tables, we observe the following:

Table 2 shows the prediction performance comparison on response time across different training densities (D1.1 to D1.5). The results indicate that the proposed R2SL model consistently achieves superior performance in both MAE and RMSE metrics. Specifically, R2SL achieves lower error rates across all densities, demonstrating its robustness and accuracy. For instance, at the lowest data density (D1.1), R2SL achieves significantly better performance compared to traditional models like UPCC and D2E-LF, which exhibit much higher error rates. When compared to QoSGNN, R2SL shows an improvement of approximately 39\% in MAE and 6\% in RMSE at the lowest data density. As the data density increases (e.g., from D1.1 to D1.5), R2SL continues to show superior performance, maintaining lower error rates and demonstrating improvements of up to 45\% in MAE and 10\% in RMSE over QoSGNN.

Table 3 presents the performance comparison on throughput across different training densities (D1.1 to D1.5). Similar to the response time results, R2SL demonstrates superior performance in both MAE and RMSE metrics. At the lowest data density (D1.1), R2SL achieves substantially better performance compared to other models, including NFMF and QoSGNN. For example, R2SL shows an improvement of approximately 38\% in MAE and 27\% in RMSE over QoSGNN. As the data density increases, R2SL consistently maintains lower MAE and RMSE, with improvements of up to 42\% in MAE and 25\% in RMSE compared to QoSGNN. This consistent performance across different densities highlights the robustness and applicability of the R2SL model in practical QoS prediction scenarios.

\vspace{-1em}
\begin{figure}[htbp]
    \centering
    \subfigure[Feature activation rate of virtual network state awareness expert]{
        \includegraphics[width=0.45\textwidth]{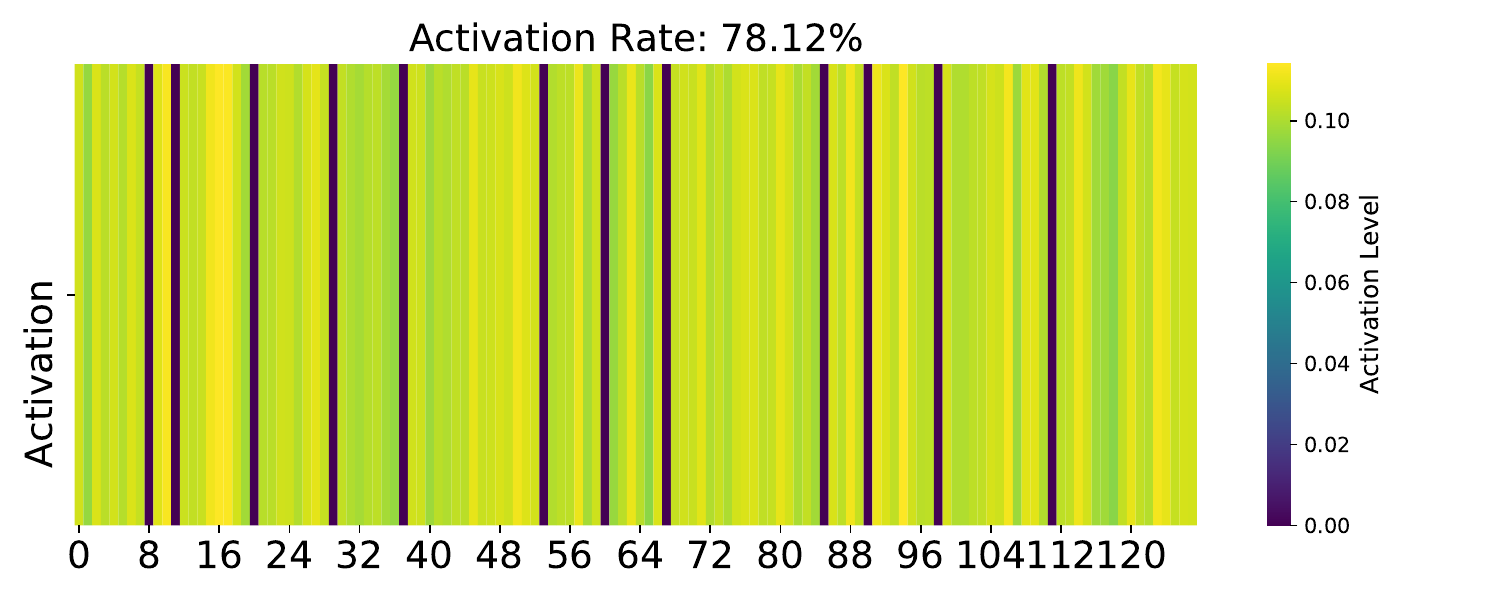}
    }
    \subfigure[Feature activation rate of physical area network state expert]{
        \includegraphics[width=0.45\textwidth]{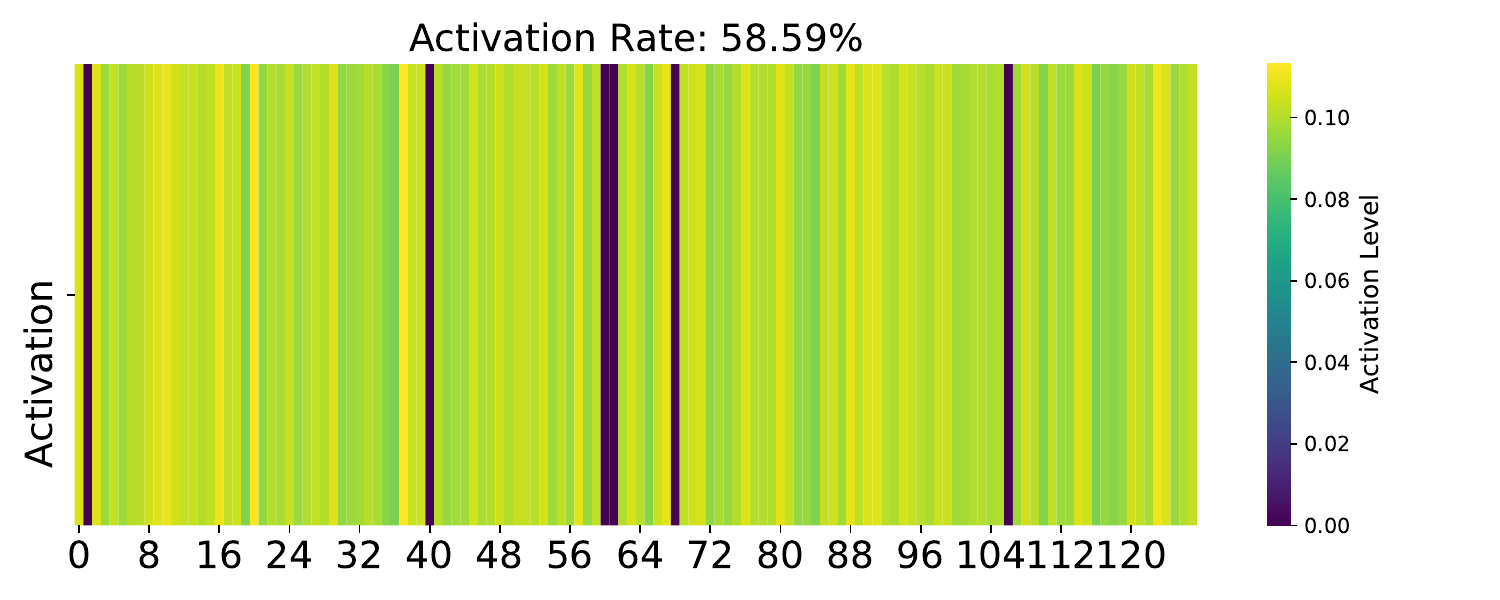}
    }
    \subfigure[Feature of gate network when user 0 accesses service 47]{
        \includegraphics[width=0.45\textwidth]{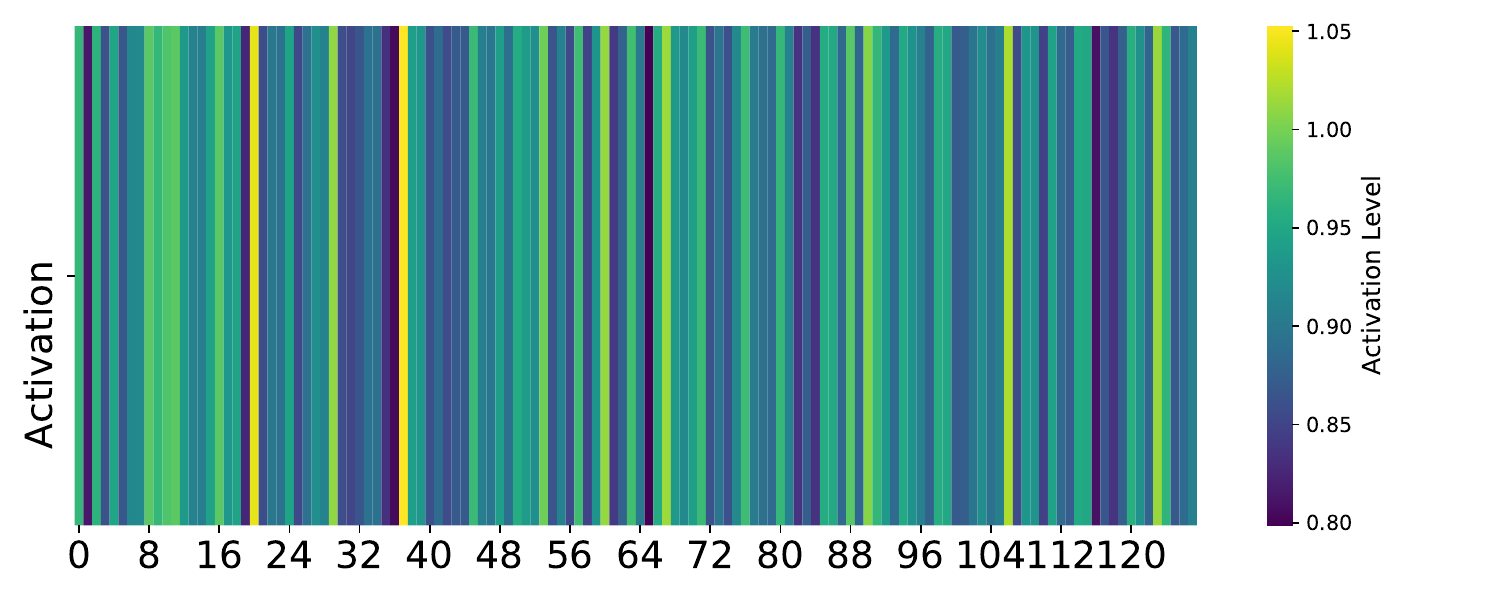}
    }
    \vspace{-1em}
    \caption{Feature activation state when user 0 accesses service 47}
    \vspace{-1em}
    \label{fig:subfigures}
\end{figure}

\subsection{ The Effect of Regional Network Latent States}

To discern the influence of network latent states on prediction performance, we instituted four comparative experiments:

a) R2SL: Employs the default state fusion map encompassing all known features and latent states.

b) R2SL W/O C: Excludes the physical region network latent states from the R2SL's feature fusion map.

c) R2SL W/O A: Omits virtual network network latent states from the R2SL's feature fusion map.

d) R2SL W/O H: Removes all latent states.

As illustrated in Fig. 8, R2SL demonstrates a prediction performance that outstrips the rest. Experiment d), by virtue of including all the network latent states found in a), b), and c), can harness a richer set of network state data for refined prediction accuracy. Additionally, d) proves especially potent in scenarios characterized by sparser training data. Contrastingly, a) which relies solely on known data points, underperforms in both MAE and RMSE metrics when set against its latent state-utilizing counterparts. These findings underscore the pivotal role latent states play as bolstering features for QoS prediction.

\begin{figure} 
\centering 

\subfigtopskip=2pt 
\subfigbottomskip=2pt 

\subfigure[MAE]{\includegraphics[width=0.495\linewidth]{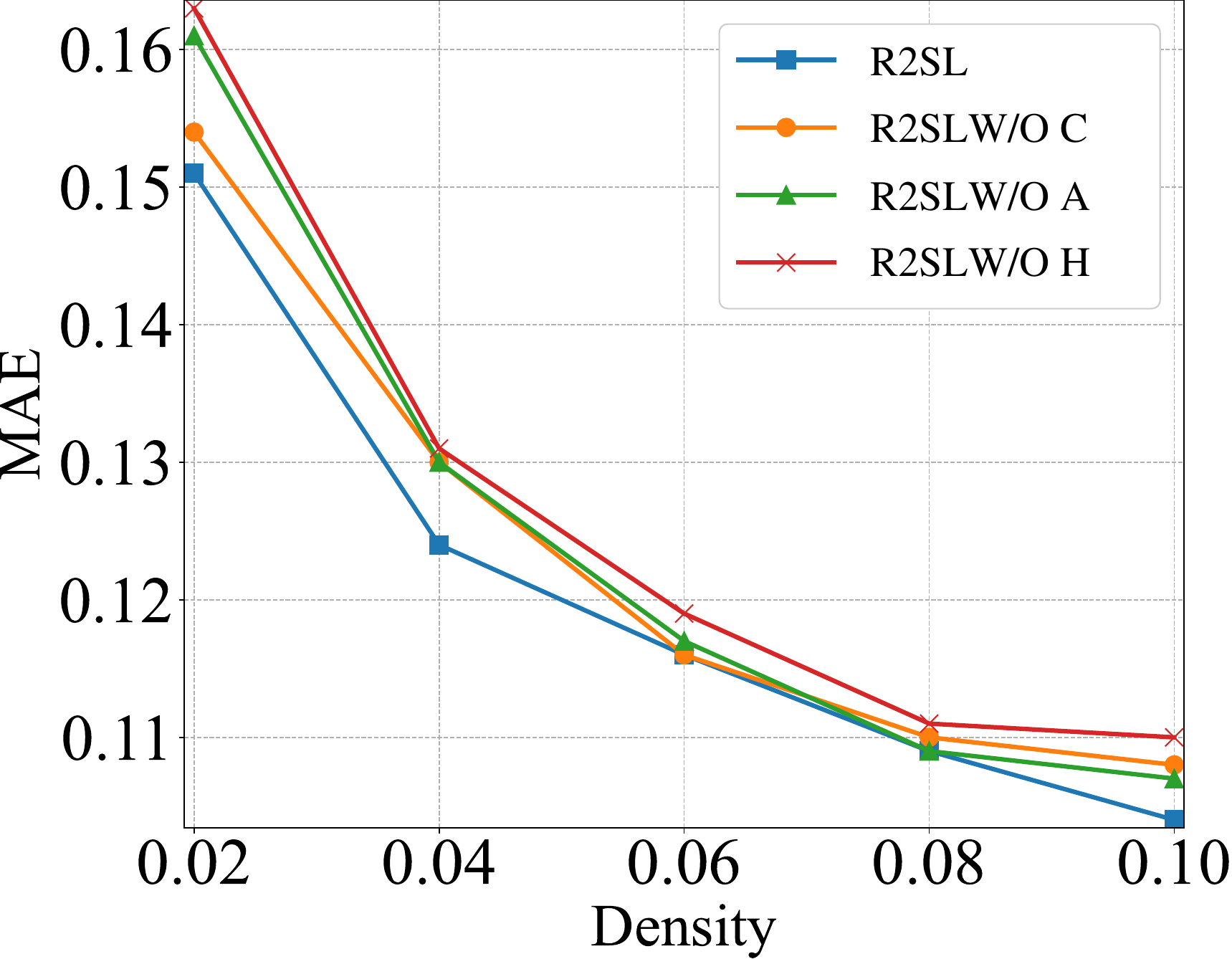}}
\subfigure[RMSE]{\includegraphics[width=0.495\linewidth]{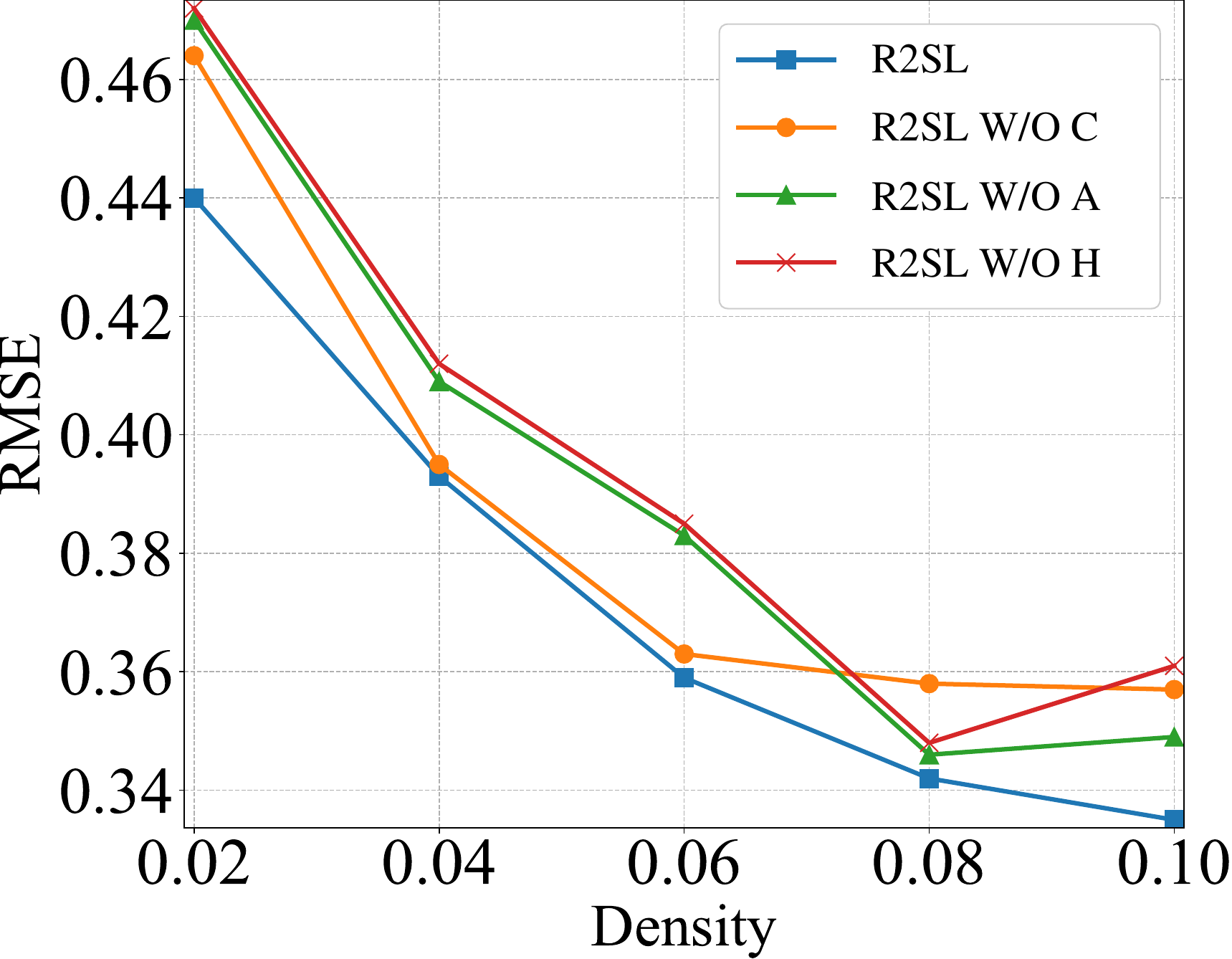}}

 \vspace{-1em}
\caption{The results for different regional network states}
 \vspace{-1em}




\end{figure}

\textbf{Latent state-aware visualization.}  Figure 7 shows a visualization of the latent feature perception for a specific access request (user 0 visits service 47).
The fig 7a suggests that virtual network state expert(VE) has a significant number of active features. This can be indicative of a strong response from the gating mechanism applied to VE latent features. 
The high activation rate mean that the network is effectively learning and utilizing the features from VE features for the prediction task. 
The activation rate of physical area network expert (PE) is lower compared to VE. This indicates that the features from PE latent features are less utilized. 
This may be due to the fact that the user and the network factors of the service are more affected by the gateway.
Figure 7c illustrates the weights of the gated network. The visualization of gate network shows the activation levels across its features. 
%
%
The gate network has a strong influence on which features are activated, indicating effective feature selection. 
The visualizations and network structure suggest that the model is effectively learning and utilizing features from different components. 
In the particular service access case, the high activation rate in virtual network latent features highlights the importance of Virtual Network-related features, while the moderate activation rate in physical area latent features shows the influence of Physical area-related features. 
%

\subsection{Loss Function for Label Imbalance}

To ascertain the efficacy of our introduced S-Huber loss function, we employed diverse loss functions on the standard R2SL network with the D1.1 dataset. Throughout the evaluation, all parameters remained consistent, defaulting to their standard values. The hyperparameter $\delta$ for the Huber loss was calibrated identically to that of the S-Huber loss, set at $\delta=0.5$. We conducted experiments using MAE, MSE, Huber loss and S-Huber loss, respectively. The MAE and MSE loss functions were derived from the standard formulations provided in TensorFlow.

AS shown in Fig.9, our findings indicated that the MSE loss trailed in effectiveness. Both MAE and Huber loss exhibited akin trajectories. 
The S-Huber loss performs well, which is the dataset with mostly short response times interspersed with prediction errors with long-tail label anomalies.
While the Huber loss manifested a diminished sensitivity to outliers compared to the MSE loss and showcased greater resilience than the MAE loss, the S-Huber loss adeptly amalgamated the strengths of both, leading in both MAE and RMSE performance metrics.

\textbf{Visualization of the training process of S-Huber Loss.}  In our study, we compared the performance of the proposed smooth Huber Loss with the traditional Huber Loss on both RT and TP datasets.
The visualization in Fig. 10 illustrates the loss values across 1000 samples for D1.1 and D2.1.
For both RT and TP datasets, the Smooth Huber Loss consistently exhibits lower loss values, particularly for extreme values, compared to the traditional Huber Loss.
This indicates that the Smooth Huber Loss is more robust and less sensitive to outliers, which helps in achieving smoother and more stable training.
The reduced impact of extreme loss values is particularly evident in the TP dataset, where the traditional Huber Loss shows significant spikes that are effectively mitigated by the Smooth Huber Loss.

This prediction result (Table 2 and 3) suggests that the smooth Huber Loss can better capture the underlying patterns of the data, leading to improved prediction performance. The reduction in extreme loss values confirms the efficacy of the smooth Huber Loss in handling outliers, thereby enhancing the overall robustness of the model.

\begin{figure} 
\centering 

\subfigtopskip=2pt 
\subfigbottomskip=2pt 

\subfigure[MAE]{\includegraphics[width=0.49\linewidth]{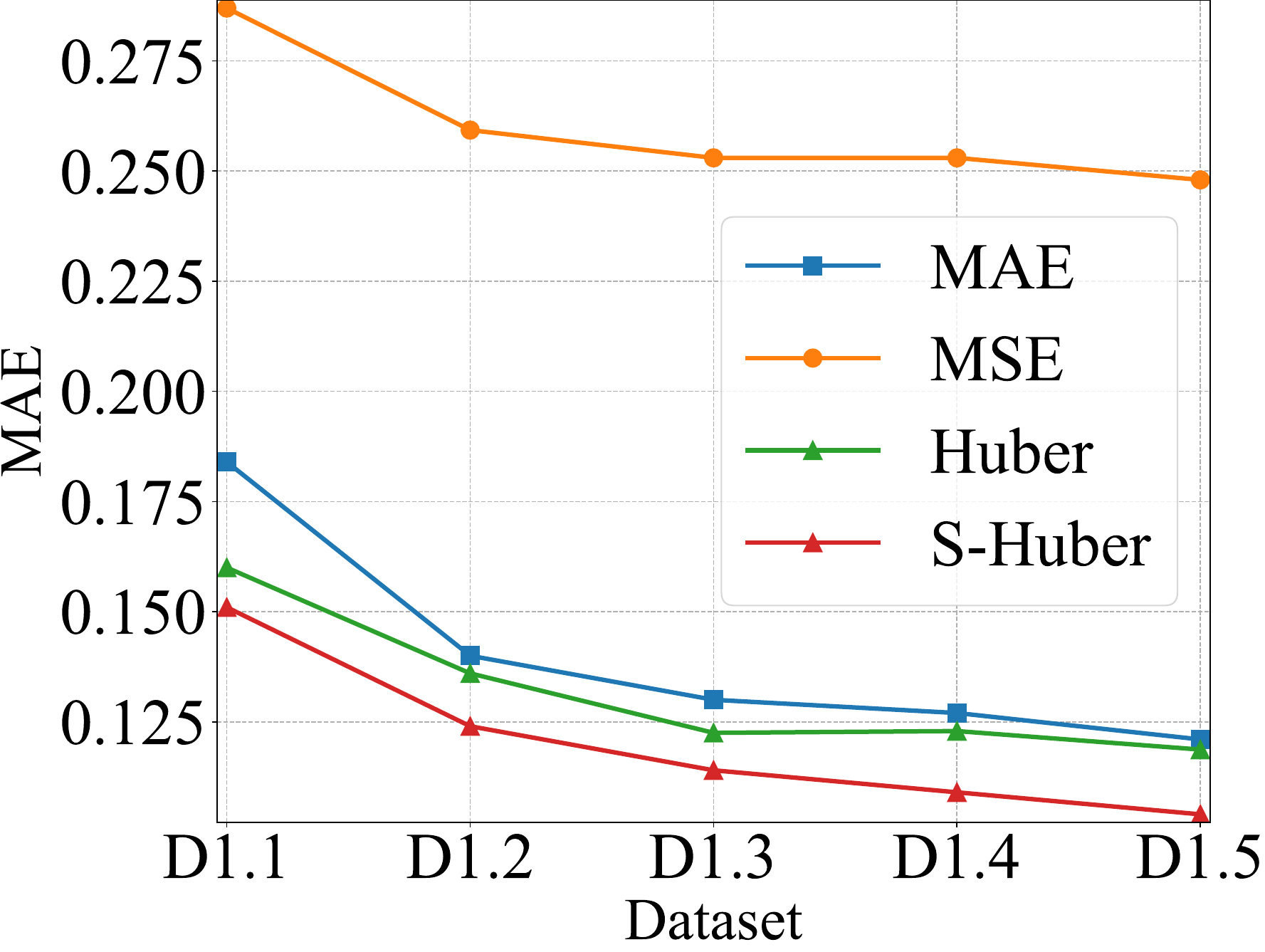}}
\subfigure[RMSE]{\includegraphics[width=0.49\linewidth]{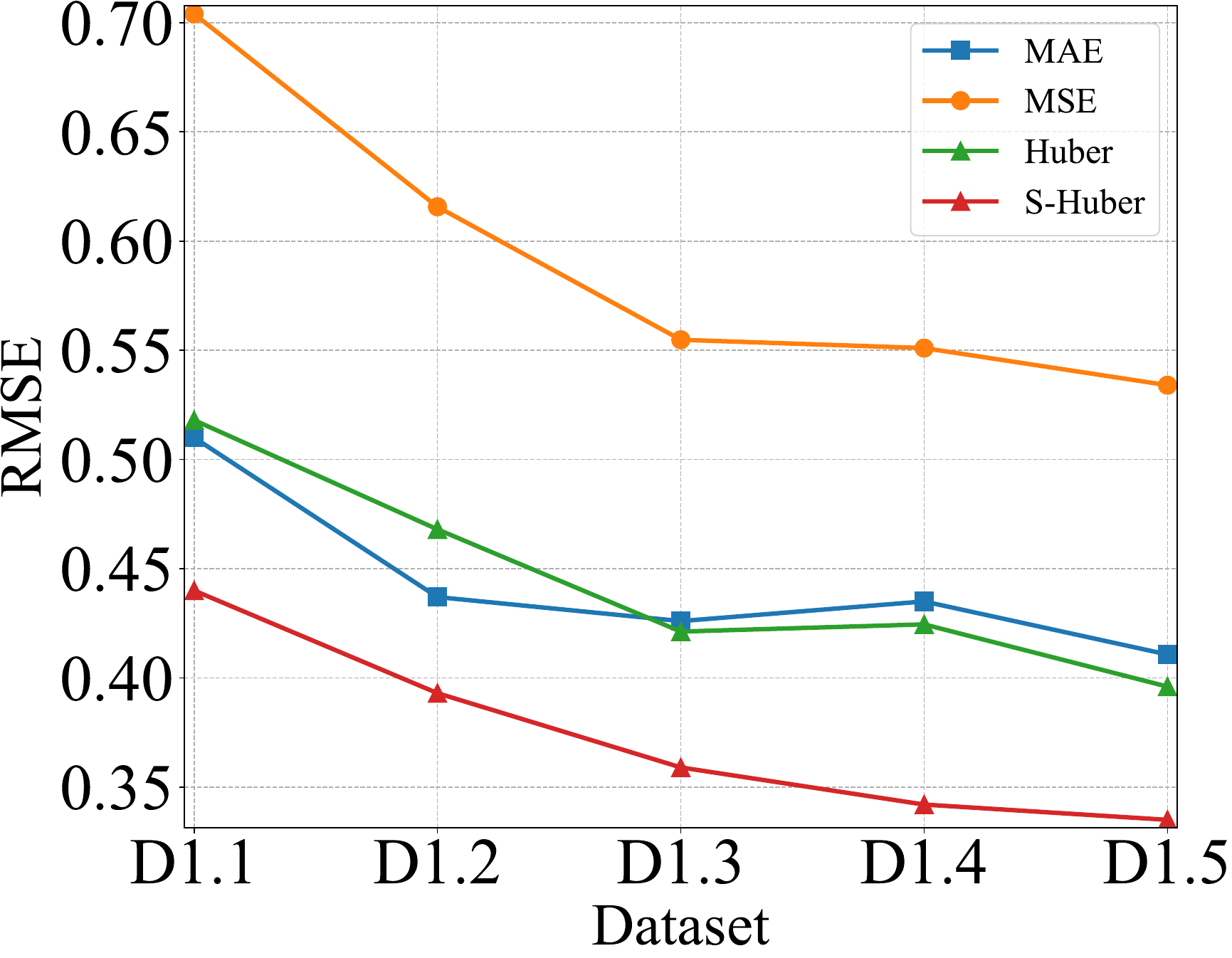}}
 \vspace{-1em}
\caption{The results of R2SL with different loss function}
\end{figure}

\begin{figure} 
\centering 

\subfigtopskip=2pt 
\subfigbottomskip=2pt 

\subfigure[The smoothness of the loss during RT prediction]{\includegraphics[width=0.99\linewidth]{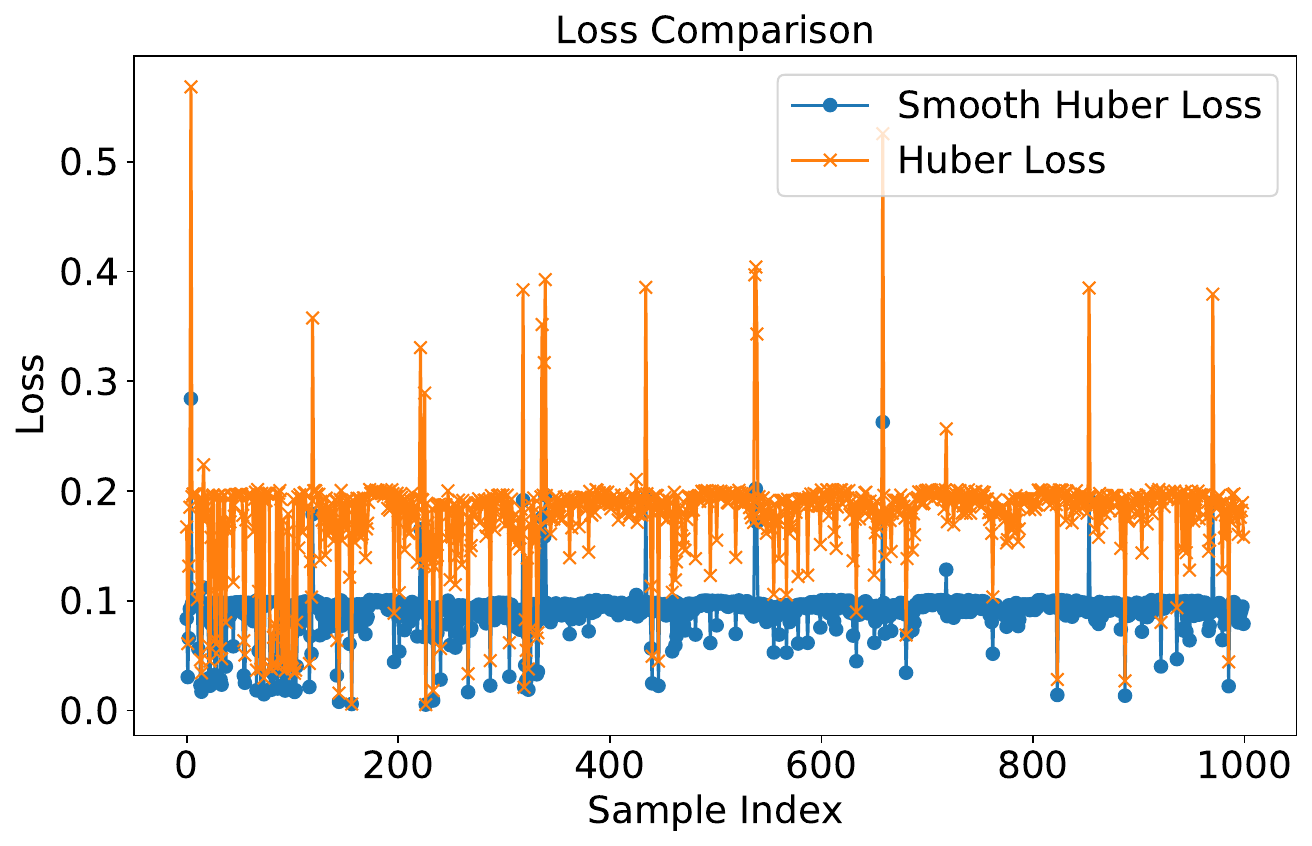}}
\subfigure[The smoothness of the loss during TP prediction]{\includegraphics[width=0.99\linewidth]{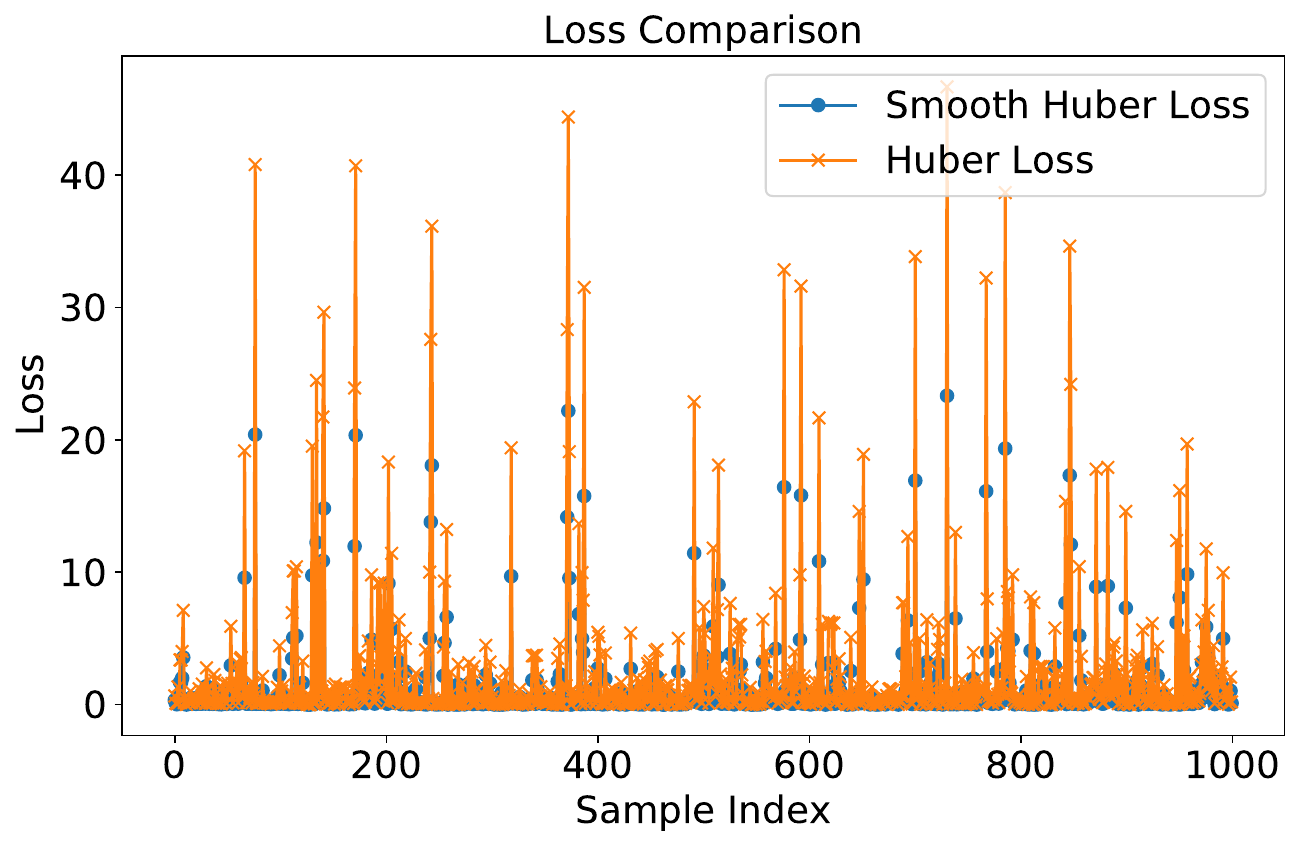}}
 \vspace{-1em}
\caption{The results of R2SL with different loss function}
 \vspace{-1em}
\end{figure}

\subsection{Ablation experiment}
In this section, we conduct ablation experiments to evaluate the effectiveness of our proposed methods. The experiments are divided into two parts: 1) comparing Sparsely Activated Mixture of Experts (MOE) with standard MOE, and 2) analyzing the impact of the parameter $\psi$ on model performance.

\subsubsection{Sparsely Activated MOE and MOE}
Fig. 11 illustrates the results of the R2SL model with sparsely activated MOE and standard MOE across different datasets (D1.1 to D1.5). From Fig. 11(a), it is evident that the sparsely activated MOE consistently outperforms the standard MOE in terms of MAE across all datasets. The MAE for the sparsely activated MOE decreases from 0.14 to 0.10 as we move from D1.1 to D1.5, demonstrating a more robust and accurate prediction capability. Similarly, Fig. 11(b) shows that the sparsely activated MOE also achieves lower RMSE values compared to the standard MOE, with RMSE decreasing from 0.46 to 0.34 across the datasets.

\begin{figure} 
\centering 

\subfigtopskip=2pt 
\subfigbottomskip=2pt 

\subfigure[MAE]{\includegraphics[width=0.49\linewidth]{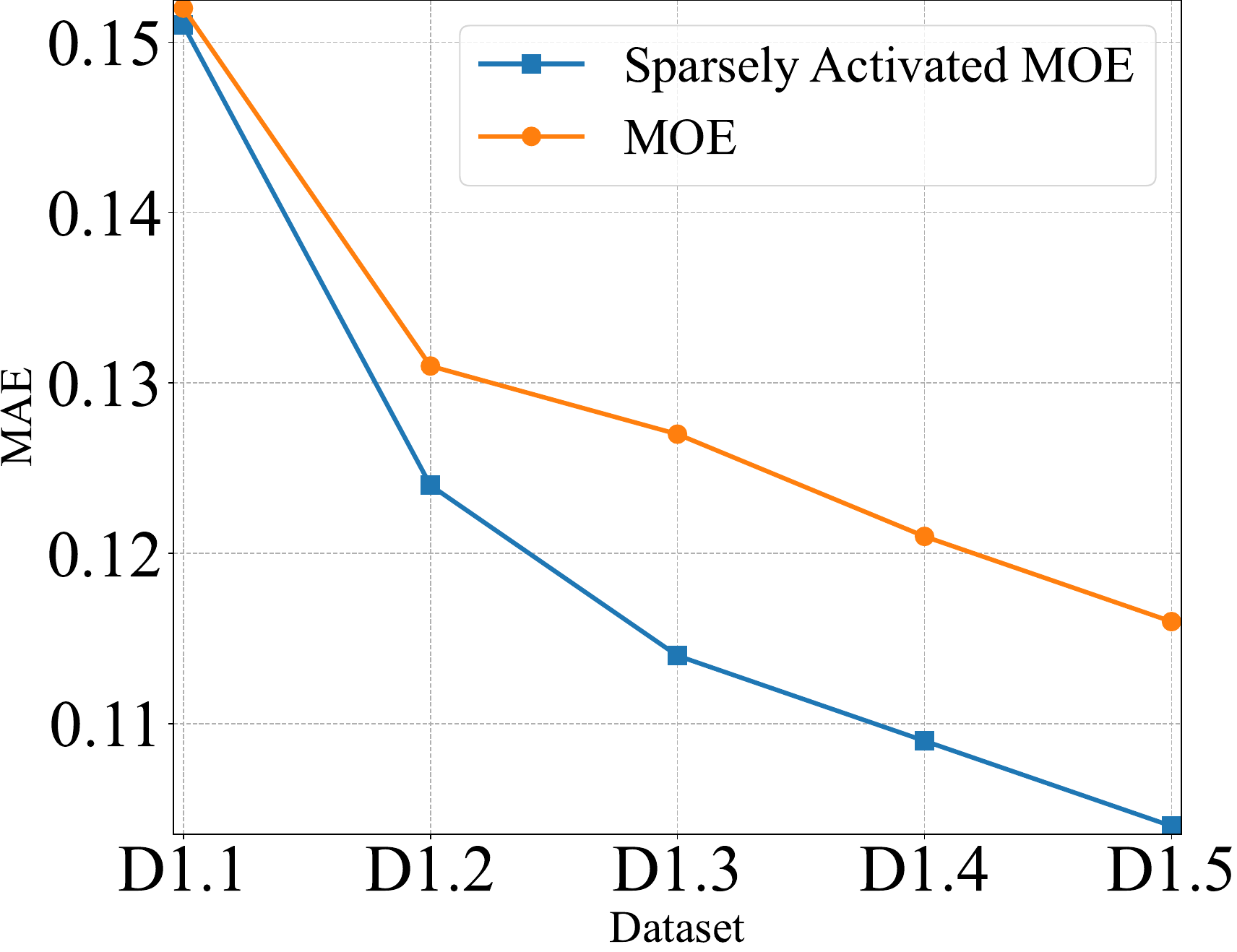}}
\subfigure[RMSE]{\includegraphics[width=0.49\linewidth]{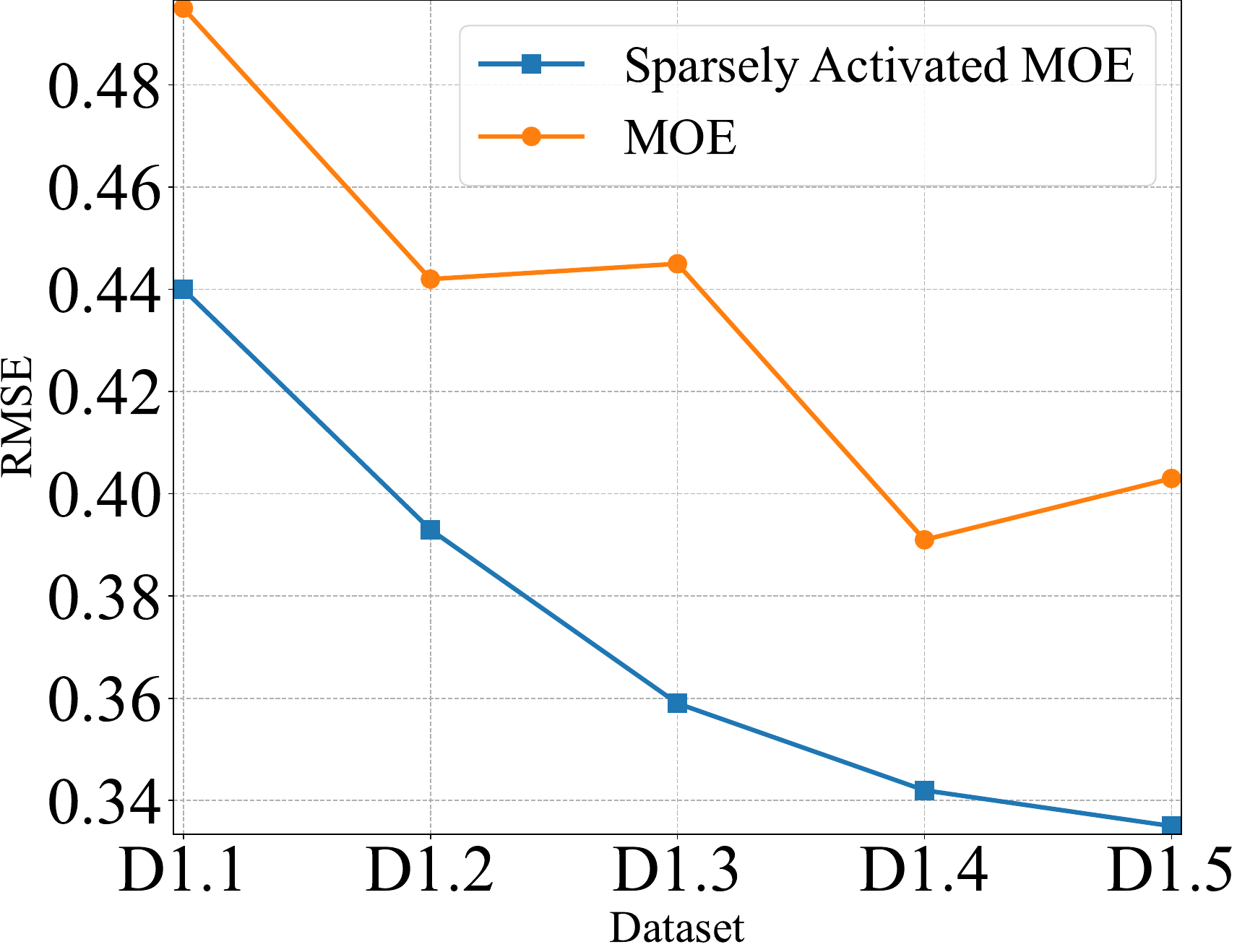}}
 \vspace{-1em}
\caption{The results of R2SL with Sparsely Activated MOE and MOE}
\end{figure}

\subsubsection{The Effect of $\psi$}

\begin{figure} 
\centering 

\subfigtopskip=2pt 
\subfigbottomskip=2pt 

\subfigure[MAE]{\includegraphics[width=0.49\linewidth]{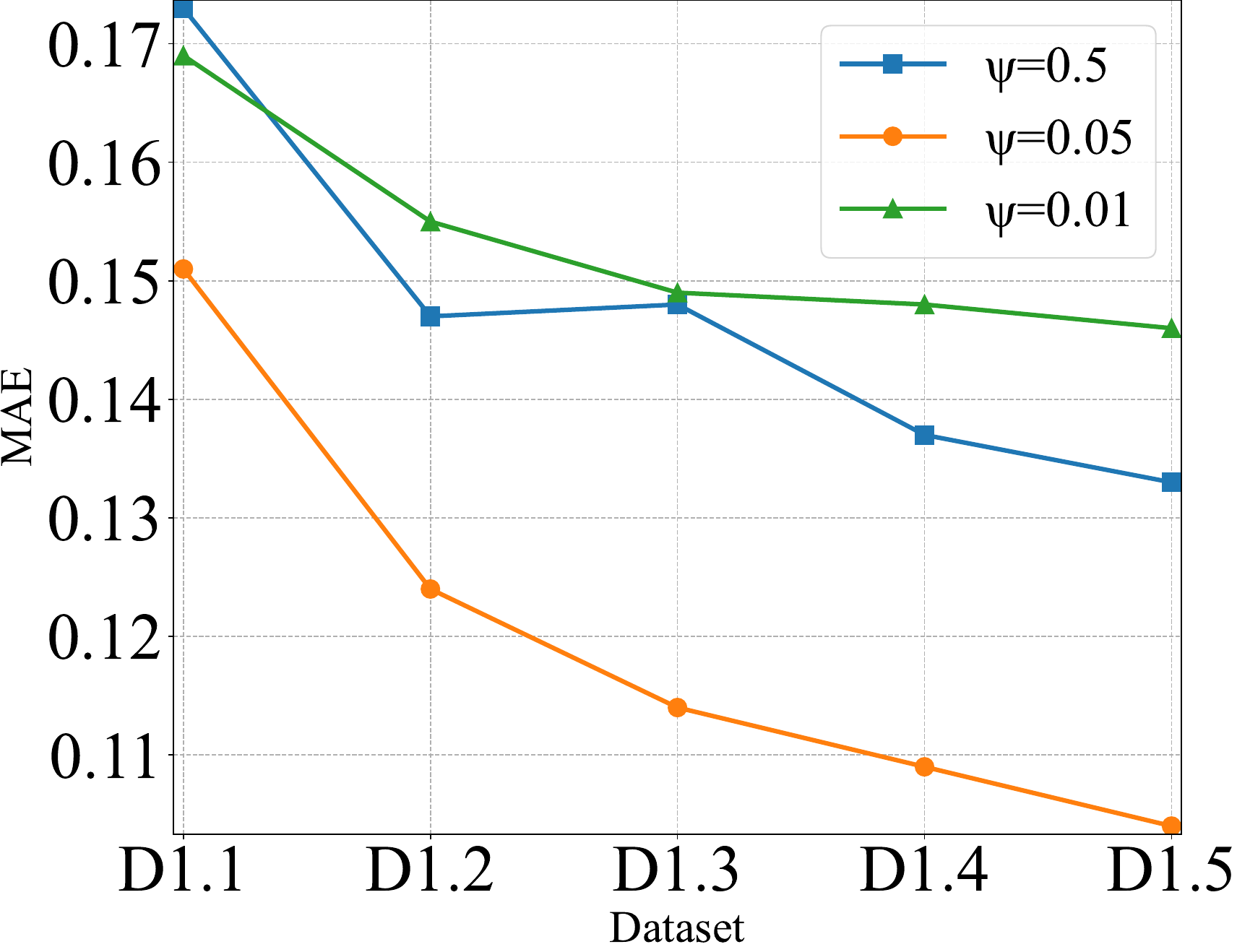}}
\subfigure[RMSE]{\includegraphics[width=0.49\linewidth]{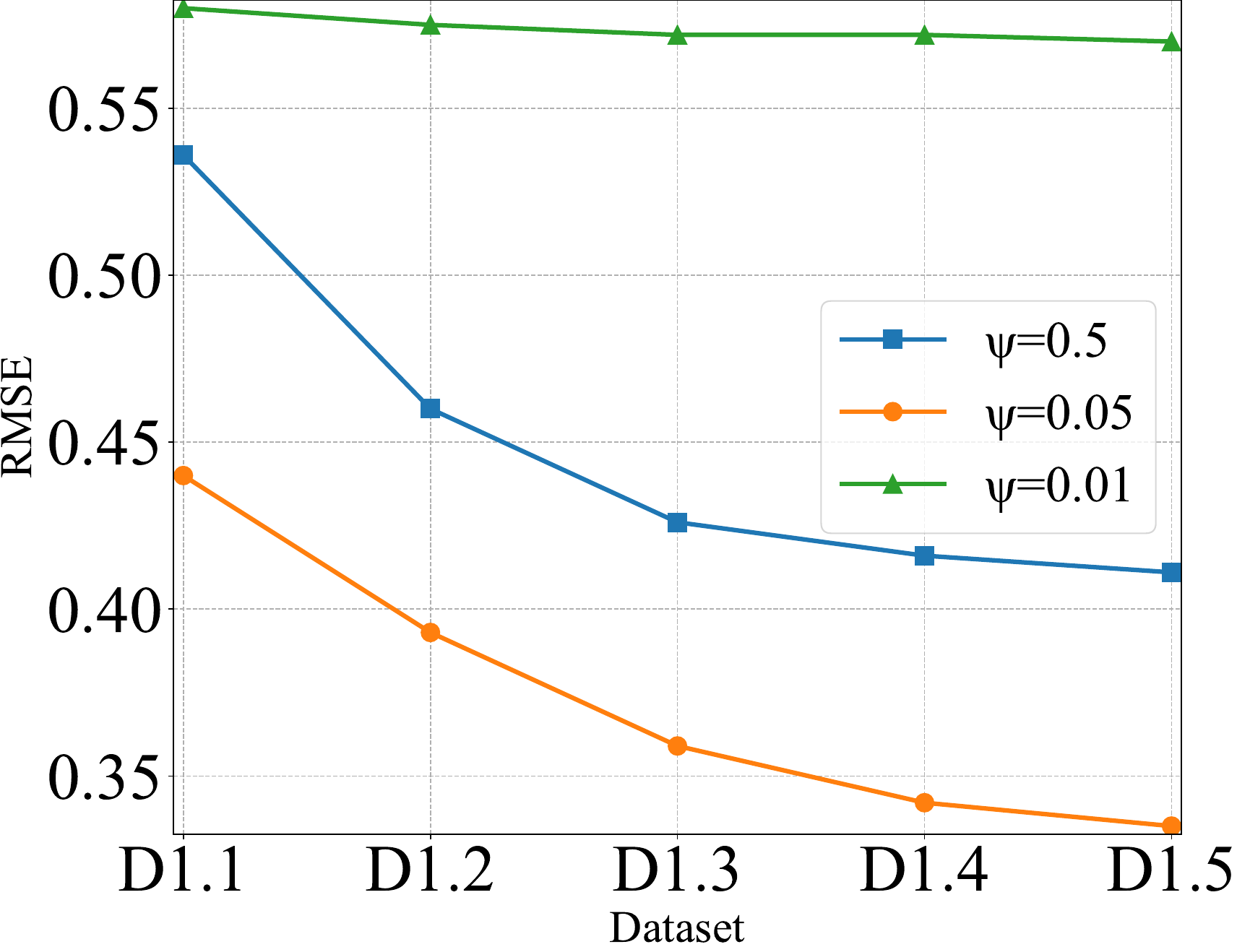}}
\caption{The results of R2SL with different  $\psi$}
 \vspace{-1em}
\end{figure}
The initial value of $\psi$ is initialized according to the historical error of QoS prediction, and the larger error value is usually 20 times of the average error, so the initial value of 0.05 is used for search.
As shown in Fig. 12(a) and Fig. 12(b), the model achieves the best performance when $\psi=0.05$. Both MAE and RMSE decrease significantly when $\psi$ is set to 0.05, indicating that this value effectively balances the trade-off between robustness and sensitivity to outliers. Lower values ($\psi=0.01$) and higher values ($\psi=0.5$) result in less optimal performance, suggesting that $\psi=0.05$ provides the right balance for the Huber loss's linear component.

\section{CONCLUSION}\label{sec:CONCLUSION}
This paper elucidates the challenges inherent in QoS prediction, pinpointing two predominant issues: data sparsity and label imbalance.
Specifically, the dearth of user data has curtailed the efficacy of prior latent factor-based prediction techniques. Concurrently, label imbalance can compromise the precision of deep models when discerning the interplay between features and QoS.
In response to these impediments, we postulate a region-centric network similarity hypothesis and put forth the Regional Network Latent State Learning Network (R2SL) model. 
And it deploys an enhanced loss function to redress label imbalance. Our empirical analyses attest to R2SL's superior performance over contemporary QoS prediction methods. 
Looking ahead, we intend to devise even more potent prediction algorithms to bolster accuracy and accommodate expansive datasets. We are also poised to weave in auxiliary contextual data to probe the influence of similarity on service caliber across geographical locales.

\ifCLASSOPTIONcaptionsoff
  \newpage
\fi

\bibliographystyle{IEEEtran}
\bibliography{IEEEabrv,my}

\begin{thebibliography}{10}
\providecommand{\url}[1]{#1}
\csname url@samestyle\endcsname
\providecommand{\newblock}{\relax}
\providecommand{\bibinfo}[2]{#2}
\providecommand{\BIBentrySTDinterwordspacing}{\spaceskip=0pt\relax}
\providecommand{\BIBentryALTinterwordstretchfactor}{4}
\providecommand{\BIBentryALTinterwordspacing}{\spaceskip=\fontdimen2\font plus
\BIBentryALTinterwordstretchfactor\fontdimen3\font minus \fontdimen4\font\relax}
\providecommand{\BIBforeignlanguage}[2]{{%
\expandafter\ifx\csname l@#1\endcsname\relax
\typeout{** WARNING: IEEEtran.bst: No hyphenation pattern has been}%
\typeout{** loaded for the language `#1'. Using the pattern for}%
\typeout{** the default language instead.}%
\else
\language=\csname l@#1\endcsname
\fi
#2}}
\providecommand{\BIBdecl}{\relax}
\BIBdecl

\bibitem{muslim2022s}
H.~S.~M. Muslim and R.~et~al., ``S-rap: relevance-aware qos prediction in web-services and user contexts,'' \emph{Knowledge and Information Systems}, vol.~64, no.~7, pp. 1997--2022, 2022.

\bibitem{zheng2020web}
Z.~Zheng, X.~Li, M.~Tang, F.~Xie, and M.~R. Lyu, ``Web service qos prediction via collaborative filtering: A survey,'' \emph{IEEE Transactions on Services Computing}, vol.~15, no.~4, pp. 2455--2472, 2020.

\bibitem{liu2019personalized}
J.~Liu and Y.~Chen, ``A personalized clustering-based and reliable trust-aware qos prediction approach for cloud service recommendation in cloud manufacturing,'' \emph{Knowledge-Based Systems}, vol. 174, pp. 43--56, 2019.

\bibitem{yao2014unified}
L.~Yao, Q.~Z. Sheng, A.~H. Ngu, J.~Yu, and A.~Segev, ``Unified collaborative and content-based web service recommendation,'' \emph{IEEE Transactions on Services Computing}, vol.~8, no.~3, pp. 453--466, 2014.

\bibitem{gavvala2019qos}
S.~K. Gavvala, C.~Jatoth, G.~Gangadharan, and R.~Buyya, ``Qos-aware cloud service composition using eagle strategy,'' \emph{Future Generation Computer Systems}, vol.~90, pp. 273--290, 2019.

\bibitem{sefati2021qos}
S.~Sefati and N.~J. Navimipour, ``A qos-aware service composition mechanism in the internet of things using a hidden-markov-model-based optimization algorithm,'' \emph{IEEE Internet of Things Journal}, vol.~8, no.~20, pp. 15\,620--15\,627, 2021.

\bibitem{47}
J.~Park, B.~Choi, C.~Lee, and D.~Han, ``Graf: a graph neural network based proactive resource allocation framework for slo-oriented microservices,'' in \emph{International Conference on emerging Networking EXperiments and Technologies}, 2021, pp. 154--167.

\bibitem{hussain2022new}
W.~Hussain, J.~M. Merig{\'o}, M.~R. Raza, and H.~Gao, ``A new qos prediction model using hybrid iowa-anfis with fuzzy c-means, subtractive clustering and grid partitioning,'' \emph{Information Sciences}, vol. 584, pp. 280--300, 2022.

\bibitem{shao2007personalized}
L.~Shao, J.~Zhang, Y.~Wei, J.~Zhao, B.~Xie, and H.~Mei, ``Personalized qos prediction forweb services via collaborative filtering,'' in \emph{Ieee international conference on web services (icws 2007)}.\hskip 1em plus 0.5em minus 0.4em\relax IEEE, 2007, pp. 439--446.

\bibitem{liu2019context}
Z.~Liu, Q.~Z. Sheng, X.~Xu, D.~Chu, and W.~E. Zhang, ``Context-aware and adaptive qos prediction for mobile edge computing services,'' \emph{IEEE Transactions on Services Computing}, vol.~15, no.~1, pp. 400--413, 2019.

\bibitem{ghafouri2020survey}
S.~H. Ghafouri, S.~M. Hashemi, and P.~C. Hung, ``A survey on web service qos prediction methods,'' \emph{IEEE Transactions on Services Computing}, vol.~15, no.~4, pp. 2439--2454, 2020.

\bibitem{zhang2018deep}
Y.~Zhang and A.~Chung, ``Deep supervision with additional labels for retinal vessel segmentation task,'' in \emph{International conference on medical image computing and computer-assisted intervention}.\hskip 1em plus 0.5em minus 0.4em\relax Springer, 2018, pp. 83--91.

\bibitem{14}
H.~Ma, I.~King, and M.~R. Lyu, ``Effective missing data prediction for collaborative filtering,'' in \emph{Proceedings of the 30th annual international ACM SIGIR conference on Research and development in information retrieval}, 2007, pp. 39--46.

\bibitem{zhang2019location}
Y.~Zhang, C.~Yin, Q.~Wu, Q.~He, and H.~Zhu, ``Location-aware deep collaborative filtering for service recommendation,'' \emph{IEEE Transactions on Systems, Man, and Cybernetics: Systems}, vol.~51, no.~6, pp. 3796--3807, 2019.

\bibitem{86}
Z.~Wang, X.~Zhang, M.~Yan, L.~Xu, and D.~Yang, ``Hsa-net: Hidden-state-aware networks for high-precision qos prediction,'' \emph{IEEE Transactions on Parallel and Distributed Systems}, vol.~33, no.~6, pp. 1421--1435, 2021.

\bibitem{carlkadie1998empirical}
J.~B.~D. CarlKadie, ``Empirical analysis of predictive algorithms for collaborative filtering,'' \emph{Microsoft Research Microsoft Corporation One Microsoft Way Redmond, WA}, vol. 98052, 1998.

\bibitem{27}
X.~Luo, M.~Zhou, Y.~Xia, Q.~Zhu, A.~C. Ammari, and A.~Alabdulwahab, ``Generating highly accurate predictions for missing qos data via aggregating nonnegative latent factor models,'' \emph{IEEE transactions on neural networks and learning systems}, vol.~27, no.~3, pp. 524--537, 2016.

\bibitem{wu2022double}
D.~Wu, P.~Zhang, Y.~He, and X.~Luo, ``A double-space and double-norm ensembled latent factor model for highly accurate web service qos prediction,'' \emph{IEEE Transactions on Services Computing}, vol.~16, no.~2, pp. 802--814, 2022.

\bibitem{xu2021nfmf}
J.~Xu, L.~Xiao, Y.~Li, M.~Huang, Z.~Zhuang, T.-H. Weng, and W.~Liang, ``Nfmf: neural fusion matrix factorisation for qos prediction in service selection,'' \emph{Connection Science}, vol.~33, no.~3, pp. 753--768, 2021.

\bibitem{lu2023feature}
T.~Lu, X.~Zhang, Z.~Wang, and M.~Yan, ``A feature distribution smoothing network based on gaussian distribution for qos prediction,'' in \emph{2023 IEEE International Conference on Web Services (ICWS)}.\hskip 1em plus 0.5em minus 0.4em\relax IEEE Computer Society, 2023, pp. 687--694.

\bibitem{3}
Z.~Zheng, H.~Ma, M.~R. Lyu, and I.~King, ``Qos-aware web service recommendation by collaborative filtering,'' \emph{IEEE Transactions on services computing}, vol.~4, no.~2, pp. 140--152, 2010.

\bibitem{zhang2021probability}
W.~Zhang, L.~Xu, M.~Yan, Z.~Wang, and C.~Fu, ``A probability distribution and location-aware resnet approach for qos prediction,'' \emph{Journal of Web Engineering}, vol.~20, no.~4, pp. 1251--1290, 2021.

\bibitem{chattopadhyay2022offdq}
S.~Chattopadhyay, R.~Chanda, S.~Kumar, and C.~Adak, ``Offdq: an offline deep learning framework for qos prediction,'' in \emph{Proceedings of the ACM Web Conference 2022}, 2022, pp. 1987--1996.

\bibitem{19}
J.~Liu, M.~Tang, Z.~Zheng, X.~Liu, and S.~Lyu, ``Location-aware and personalized collaborative filtering for web service recommendation,'' \emph{IEEE Transactions on Services Computing}, vol.~9, no.~5, pp. 686--699, 2015.

\bibitem{wu2020data}
D.~Wu, X.~Luo, and M.~e.~a. Shang, ``A data-characteristic-aware latent factor model for web services qos prediction,'' \emph{IEEE Transactions on Knowledge and Data Engineering}, vol.~34, no.~6, pp. 2525--2538, 2020.

\bibitem{chowdhury2020cahphf}
R.~R. Chowdhury, S.~Chattopadhyay, and C.~Adak, ``Cahphf: context-aware hierarchical qos prediction with hybrid filtering,'' \emph{IEEE Transactions on Services Computing}, vol.~15, no.~4, pp. 2232--2247, 2020.

\bibitem{liang2021recurrent}
T.~Liang and C.~et~al., ``Recurrent neural network based collaborative filtering for qos prediction in iov,'' \emph{IEEE Transactions on Intelligent Transportation Systems}, vol.~23, no.~3, pp. 2400--2410, 2021.

\bibitem{li2021topology}
J.~Li, H.~Wu, J.~Chen, Q.~He, and C.-H. Hsu, ``Topology-aware neural model for highly accurate qos prediction,'' \emph{IEEE Transactions on Parallel and Distributed Systems}, vol.~33, no.~7, pp. 1538--1552, 2021.

\bibitem{xia2021joint}
Y.~Xia, D.~Ding, Z.~Chang, and F.~Li, ``Joint deep networks based multi-source feature learning for qos prediction,'' \emph{IEEE Transactions on Services Computing}, vol.~15, no.~4, pp. 2314--2327, 2021.

\bibitem{1}
K.~Lee, J.~Park, and J.~Baik, ``Location-based web service qos prediction via preference propagation for improving cold start problem,'' in \emph{IEEE International Conference on Web Services}.\hskip 1em plus 0.5em minus 0.4em\relax IEEE, 2015, pp. 177--184.

\bibitem{2}
X.~Chen, X.~Liu, Z.~Huang, and H.~Sun, ``Regionknn: A scalable hybrid collaborative filtering algorithm for personalized web service recommendation,'' in \emph{IEEE international conference on web services}.\hskip 1em plus 0.5em minus 0.4em\relax IEEE, 2010, pp. 9--16.

\bibitem{4}
Z.~Chen, L.~Shen, F.~Li, and D.~You, ``Your neighbors alleviate cold-start: On geographical neighborhood influence to collaborative web service qos prediction,'' \emph{Knowledge-Based Systems}, vol. 138, pp. 188--201, 2017.

\bibitem{13}
B.~Sarwar, G.~Karypis, J.~Konstan, and J.~Riedl, ``Item-based collaborative filtering recommendation algorithms,'' in \emph{Proceedings of the 10th international conference on World Wide Web}, 2001, pp. 285--295.

\bibitem{12}
Z.~Tan and L.~He, ``An efficient similarity measure for user-based collaborative filtering recommender systems inspired by the physical resonance principle,'' \emph{IEEE Access}, vol.~5, pp. 27\,211--27\,228, 2017.

\bibitem{79}
J.~A. Konstan, B.~N. Miller, D.~Maltz, J.~L. Herlocker, L.~R. Gordon, and J.~Riedl, ``Grouplens: Applying collaborative filtering to usenet news,'' \emph{Communications of the ACM}, vol.~40, no.~3, pp. 77--87, 1997.

\bibitem{8}
X.~Luo, M.~Zhou, S.~Li, Z.~You, Y.~Xia, and Q.~Zhu, ``A nonnegative latent factor model for large-scale sparse matrices in recommender systems via alternating direction method,'' \emph{IEEE transactions on neural networks and learning systems}, vol.~27, no.~3, pp. 579--592, 2015.

\bibitem{9}
Y.~Shi, M.~Larson, and A.~Hanjalic, ``Collaborative filtering beyond the user-item matrix: A survey of the state of the art and future challenges,'' \emph{ACM Computing Surveys (CSUR)}, vol.~47, no.~1, pp. 1--45, 2014.

\bibitem{35}
Y.~Zhang, Z.~Zheng, and M.~R. Lyu, ``Wspred: A time-aware personalized qos prediction framework for web services,'' in \emph{IEEE International Symposium on Software Reliability Engineering}.\hskip 1em plus 0.5em minus 0.4em\relax IEEE, 2011, pp. 210--219.

\bibitem{81}
S.~Wang, Y.~Zhao, L.~Huang, J.~Xu, and C.-H. Hsu, ``Qos prediction for service recommendations in mobile edge computing,'' \emph{Journal of Parallel and Distributed Computing}, vol. 127, pp. 134--144, 2019.

\bibitem{29}
Z.~Luo, L.~Liu, and Y.~et~al., ``Latent ability model: A generative probabilistic learning framework for workforce analytics,'' \emph{IEEE Transactions on Knowledge and Data Engineering}, vol.~31, no.~5, pp. 923--937, 2018.

\bibitem{20}
X.~He, L.~Liao, H.~Zhang, L.~Nie, X.~Hu, and T.-S. Chua, ``Neural collaborative filtering,'' in \emph{Proceedings of the 26th international conference on world wide web}, 2017, pp. 173--182.

\bibitem{83}
Q.~Zhou, H.~Wu, K.~Yue, and C.-H. Hsu, ``Spatio-temporal context-aware collaborative qos prediction,'' \emph{Future Generation Computer Systems}, vol. 100, pp. 46--57, 2019.

\bibitem{wang2016online}
H.~Wang, L.~Wang, Q.~Yu, Z.~Zheng, A.~Bouguettaya, and M.~R. Lyu, ``Online reliability prediction via motifs-based dynamic bayesian networks for service-oriented systems,'' \emph{IEEE Transactions on Software Engineering}, vol.~43, no.~6, pp. 556--579, 2016.

\bibitem{xiong2018deep}
R.~Xiong, J.~Wang, N.~Zhang, and Y.~Ma, ``Deep hybrid collaborative filtering for web service recommendation,'' \emph{Expert systems with Applications}, vol. 110, pp. 191--205, 2018.

\bibitem{liu2023qosgnn}
M.~Liu, H.~Xu, Q.~Z. Sheng, and Z.~Wang, ``Qosgnn: Boosting qos prediction performance with graph neural networks,'' \emph{IEEE Transactions on Services Computing}, 2023.

\bibitem{77}
F.~Ye, Z.~Lin, C.~Chen, Z.~Zheng, and H.~Huang, ``Outlier-resilient web service qos prediction,'' in \emph{Proceedings of the Web Conference 2021}, 2021, pp. 3099--3110.

\bibitem{21}
Y.~Zhang, C.~Yin, Q.~Wu, Q.~He, and H.~Zhu, ``Location-aware deep collaborative filtering for service recommendation,'' \emph{IEEE Transactions on Systems, Man, and Cybernetics: Systems}, 2019.

\bibitem{zou2022ncrl}
G.~Zou, S.~Wu, S.~Hu, C.~Cao, Y.~Gan, B.~Zhang, and Y.~Chen, ``Ncrl: Neighborhood-based collaborative residual learning for adaptive qos prediction,'' \emph{IEEE Transactions on Services Computing}, 2022.

\end{thebibliography}

\end{document}